\documentclass[acmtog]{acmart}
\acmSubmissionID{173}

\usepackage{booktabs} 

\acmJournal{TOG}

\setcopyright{acmcopyright}
\acmJournal{TOG}
\acmYear{2021}
\acmVolume{40}
\acmNumber{4}
\acmArticle{149}
\acmMonth{8} 

\acmDOI{10.1145/3450626.3459756}

\newcommand\mR {\mathcal{R}} 

\begin{document}

\title{Editable Free-Viewpoint Video using a Layered Neural Representation}

\author{Jiakai Zhang}
\orcid{0000-0001-9477-3159}
\affiliation{%
  \institution{ShanghaiTech University}
  \city{Shanghai}
  \country{China}}
\affiliation{%
	\institution{Stereye Intelligent Technology Co.,Ltd.}
	\country{China}}
\email{zhangjk@shanghaitech.edu.cn}

\author{Xinhang Liu}
\affiliation{%
  \institution{ShanghaiTech University}
  \city{Shanghai}
  \country{China}
}
\email{liuxh2@shanghaitech.edu.cn}

\author{Xinyi Ye}
\affiliation{%
	\institution{ShanghaiTech University}
	\city{Shanghai}
	\country{China}
}
\email{yexy2@shanghaitech.edu.cn}

\author{Fuqiang Zhao}
\affiliation{%
	\institution{ShanghaiTech University}
	\city{Shanghai}
	\country{China}
}
\email{zhaofq@shanghaitech.edu.cn}

\author{Yanshun Zhang}
\affiliation{%
	\institution{DGene Digital Technology Co., Ltd.}
	\country{China}
}
\email{yanshun.zhang@dgene.com}

\author{Minye Wu}
\affiliation{%
	\institution{ShanghaiTech University}
	\city{Shanghai}
	\country{China}
}
\email{wumy@shanghaitech.edu.cn}

\author{Yingliang Zhang}
\affiliation{%
	\institution{DGene Digital Technology Co., Ltd.}
	\country{China}
}
\email{yingliang.zhang@dgene.com}

\author{Lan Xu}
\affiliation{%
	\institution{ShanghaiTech University}
	\city{Shanghai}
	\country{China}
}
\email{xulan1@shanghaitech.edu.cn}
\authornote{The corresponding authors are Lan Xu (xulan1@shanghaitech.edu.cn) and Jingyi Yu (yujingyi@shanghaitech.edu.cn). }

\author{Jingyi Yu}
\affiliation{%
	\institution{ShanghaiTech University}
	\city{Shanghai}
	\country{China}
}
\email{yujingyi@shanghaitech.edu.cn}
\authornotemark[1]

\begin{abstract}
Generating free-viewpoint videos is critical for immersive VR/AR experience, but recent neural advances still lack the editing ability to manipulate the visual perception for large dynamic scenes.
To fill this gap, in this paper, we propose the first approach for editable free-viewpoint video generation for large-scale view-dependent dynamic scenes using only 16 cameras.
The core of our approach is a new layered neural representation, where each dynamic entity, including the environment itself, is formulated into a spatio-temporal coherent neural layered radiance representation called ST-NeRF.
Such a layered representation supports manipulations of the dynamic scene while still supporting a wide free viewing experience.
In our ST-NeRF, we represent the dynamic entity/layer as a continuous function, which achieves the disentanglement of location, deformation as well as the appearance of the dynamic entity in a continuous and self-supervised manner.
We propose a scene parsing 4D label map tracking to disentangle the spatial information explicitly and a continuous deform module to disentangle the temporal motion implicitly.
An object-aware volume rendering scheme is further introduced for the re-assembling of all the neural layers.
We adopt a novel layered loss and motion-aware ray sampling strategy to enable efficient training for a large dynamic scene with multiple performers, 
Our framework further enables a variety of editing functions, i.e., manipulating the scale and location, duplicating or retiming individual neural layers to create numerous visual effects while preserving high realism.
Extensive experiments demonstrate the effectiveness of our approach to achieve high-quality, photo-realistic, and editable free-viewpoint video generation for dynamic scenes.
\end{abstract}

%
%


%
%


\begin{teaserfigure}
  \centering
  \includegraphics[width=0.95\linewidth]{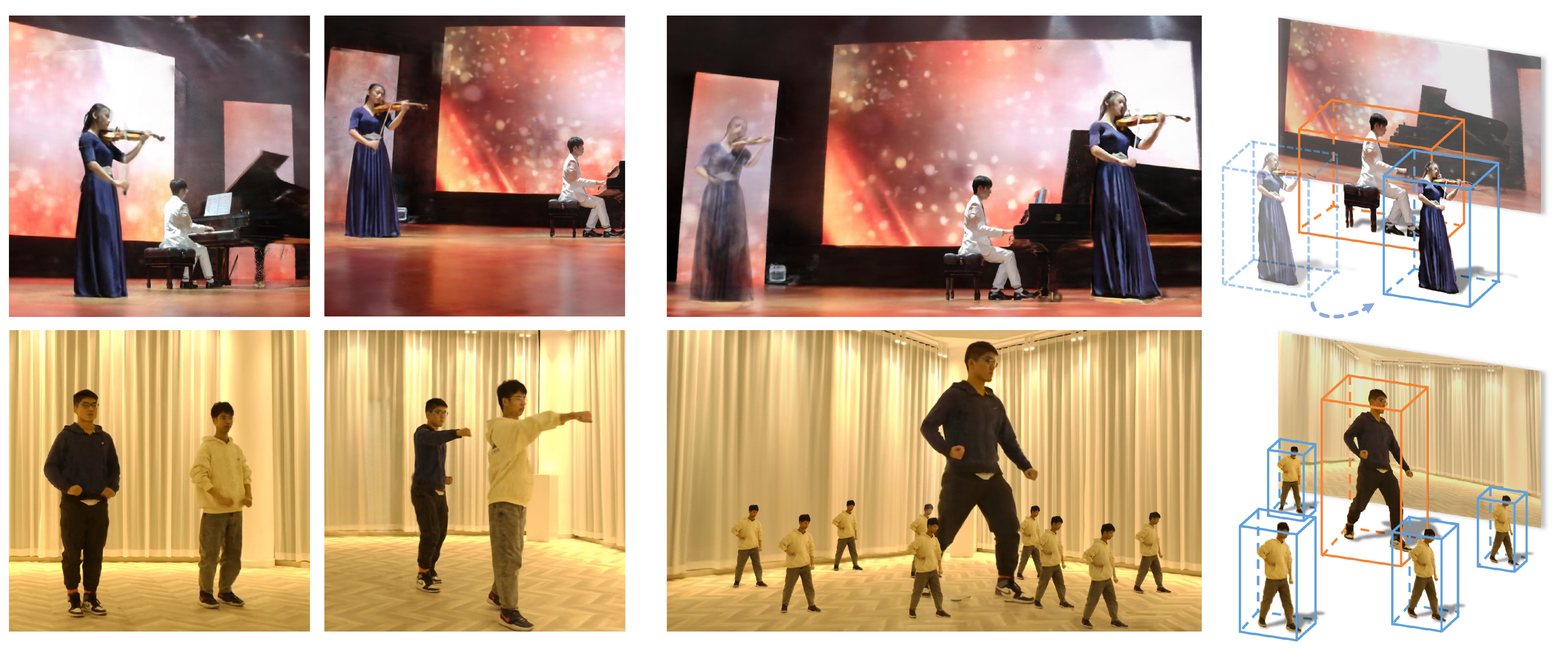}
  \caption{Our approach generates photo-realistic and editable free-viewpoint videos for dynamic scenes using a layered neural representation from 16 RGB streams. Our framework enables various editing functions, i.e., manipulating the scale and location, duplicating, adjusting transparency, or retiming for individual neural layers while supporting space-time viewing experience. From left to right in each row: two rendering results of different viewpoints without editing, the edited results in a novel viewpoint, and the corresponding 3D illustration.}
  \label{fig:teaser}
\end{teaserfigure}

\maketitle

\section{INTRODUCTION}
Novel view synthesis, one of the core tasks in computer vision and graphics, provides unique viewing experiences and has been widely used in visual effects from gaming to education, from art to entertainment.
One of the most famous examples is the “bullet-time” effects presented in the feature film The Matrix, which creates the stopping-time illusion with smooth transitions of viewpoints surrounding the actor.
Such novel view synthesis further evolves as a cutting-edge yet bottleneck technique with the rise of Virtual Reality (VR) and Augmented Reality (AR) over the last decade, which presents information in an immersive way unthinkable before.
How to generate editable free-viewpoint videos of dynamic scenes for fully natural and controllable viewing experiences in VR/AR remains unsolved and has recently attracted the substantive attention of both the computer vision and graphics communities.

For free-viewpoint video generation of human-centric dynamic scenes, early model-based solutions~\cite{Mustafa_2016_CVPR,collet2015high} rely on multi-view dome-based setup for high-fidelity reconstruction and texture rendering in novel views.
However, they are restricted by the limited reconstructed mesh resolution and suffer from the uncanny texturing output, especially for large-scale captured scenes.
On the other hand, traditional image-based rendering (IBR) techniques~\cite{gortler1996lumigraph,carranza2003free,zitnick2004high} interpolate textures in novel views directly from the dense captured viewpoints.
The free-view results, however, are vulnerable to occlusions and suffer from limited view interpolation along with the dense captured views, leading to uncanny texture details due to view blending.
The recent neural rendering techniques~\cite{NR_survey} bring huge potential for compelling photo-realistic free-viewpoint video generation via neural view blending~\cite{NRWird_CVPR19,Thies2020Image-guided} or neural scene modeling~\cite{Wu_2020_CVPR,mildenhall2020nerf, bemana2020x}. Open4D~\cite{bansal20204d} generates a free-viewpoint video enabling occlusion removal and time-freezing effects using around 15 mobile cameras, which is similar to our work.
Such data-driven approaches get rid of the heavy reliance on reconstruction accuracy or the extremely dense capture setting.
Recent work~\cite{park2020deformable,tretschk2020non,rebain2020derf,ost2020neural} extend the NeRF approach~\cite{mildenhall2020nerf} into the dynamic setting. 
However, the above solutions for dynamic scene free-viewpoint synthesis still suffer from limited capture volume or fragile human motions.
More importantly, they focus on reconstruction only, without any editing functions for the visual effects that can change the perception of the dynamic scenes.
Recently, the work~\cite{lu2020layered} enables the neural retiming effects of human motions using a monocular video.
However, it is limited in the temporal effect only, without exploring the rich 3D spatial editing functions for novel view synthesis in free-viewpoint videos. 

In this paper, we address the above challenges and present the first approach to generate editable photo-realistic free-viewpoint videos of large-scale dynamic scenes, using only 16 cameras to cover a view range up to 180 degrees.
As illustrated in Fig.~\ref{fig:teaser}, our approach marries the free-viewpoint videos with a new neural layered space-time representation.
It enables various spatial and temporal editing functions for numerous photo-realistic visual effects whilst still supporting free viewing in a wide range.

Generating such free-viewpoint videos with the photo-realistically editable new effect from a much sparser camera setting than traditional systems is non-trivial.
Our key idea is to model the space-time correlations of all the dynamic entities, including the environment itself, into a consistent neural representation so as to fully support perception and realistic manipulation of the dynamic scene.
To this end, based on the input multi-viewpoint videos, we first adopt an effective scene parsing stage to generate the coarse space-time 4D label maps of all the dynamic performers in the captured scene.
Our scene parsing utilizes the multi-view geometry prior as well as label-level cross-view tracking so as to provide an initial high-level layer-wised perception of the dynamic scene.
Then, as the core of our approach, a new neural layered representation is proposed. Each entity is formulated as a separated continuous function of both space and time, forming a spatially and temporally consistent neural radiance field (ST-NeRF) to support various editing functions. 
In our ST-NeRF for a dynamic entity, a continuous deform module is introduced to encode the temporal motion information of each dynamic entity. At the same time, the corresponding 4D label map with bounding-box from scene parsing serves as a spatial anchor to fuse the appearance information across views and timestamps. 
A novel layered loss and a motion-aware ray-sampling strategy are further adapted to enable the efficient training of our ST-NeRF of a large dynamic scene with multiple dynamic performers, as well as an object-aware volume rendering for the re-assembling of all the neural layers. Notably, when the performers interact closely, it is hard to give the correct decomposition results due to the bounding boxes of performers are highly overlapped.
%
After training our ST-NeRF, during the inference, our layered representation and explicit space-time disentanglement enable a variety of editing functions upon each dynamic entity so as to generate photo-realistic editable free-viewpoint results.
Our neural editing includes the basic operation of manipulating the input position and timing of the continuous representation of each dynamic entity.  
Thus, various spatial editing like affine transform or duplication as well as temporal editing like retiming performers' movements can be applied to each layer/entity in a depth-aware and photo-realistic manner.
To summarize, our main contributions include:
\begin{itemize} 
	\setlength\itemsep{0em}
	\item We demonstrate the new capability of editable free-viewpoint video generations from only 16 cameras, which enables various photo-realistic space-time visual editing effects whilst still supporting wide-range free viewing.
	
	\item We introduce a novel neural layered representation for large-scale dynamic scene modeling and manipulation, enabled by the
	disentanglement of location, deformation as well as the appearance of all the dynamic entities.
	
	\item We propose a layer-wise 4D label map tracking to disentangle the spatial information explicitly, as well as a continuous deform module to disentangle the temporal motion implicitly. 
	
	\item We propose a novel layered loss and a motion-aware ray-sampling strategy to efficiently train our layered neural representation for a large-scale dynamic scene with multiple performers. 
	 
\end{itemize}

\section{RELATED WORK}
\hspace*{\fill} \\
\noindent{\bf Image-based Rendering without Deep Learning.}
Traditional image-based rendering (IBR) approaches interpolate textures in novel views directly from a set of input images~\cite{chen1993view,matusik2000image,carranza2003free,zitnick2004high}.
Many methods~\cite{buehler2001unstructured,chaurasia2013depth,debevec1996modeling,hedman2017casual,snavely2006photo} rely on building an explicit 3D scene geometry first for rendering in novel views.
The method~\cite{hedman2016scalable} further utilizes an RGBD sensor to enable fast rendering. 
However, the reliance on explicit geometry makes it difficult to apply these methods to a large-scale dynamic scene. 
On the other hand, light field rendering methods~\cite{levoy1996light,gortler1996lumigraph} synthesize novel views only using implicit soft geometry representations derived from densely sampled images.
As the representative techniques, the light field rendering~\cite{levoy1996light} synthesizes novel views by filtering and interpolating view samples while the lumigraph~\cite{gortler1996lumigraph} applies coarse geometry to compensate for non-uniform sampling.
Numerous other works~\cite{davis2012unstructured,penner2017soft} explore the special structure of light fields to improve the rendering quality.
Another direction is using the multiplane images (MPIs) as 3D representation, which have been applied to model complex scene appearance~\cite{broxton2020immersive,choi2019extreme,mildenhall2019local,srinivasan2019pushing}.
However, these approaches above still cannot provide a wide-range free-viewing of a large dynamic scene, let alone editing various dynamic entities. 
To model dynamic scenes, the prior works~\cite{lipski2010virtual,zitnick2004high,carranza2003free,bansal20204d} require multi-view, time-synchronized videos as input for rendering various space-time visual effects.
Zitnick~\emph{et al.}~\cite{zitnick2004high} use depth maps estimated from multi-view stereo to guide viewpoint interpolation. 
Carranza~\emph{et al.}~\cite{carranza2003free} uses a multi-view system to recover 3D models from silhouettes for synthesizing novel views from arbitrary perspectives. 
Kumar~\emph{et al.}~\cite{kumar2019superpixel} uses a monocular video as input to recover a dynamic scene by a conventional optimization approach with two assumptions of scene and deformation, which is 1) The dynamic scene can be approximated by multiple piece-wise planar surfaces with rigid motions, 2) the deformation of the scene is locally rigid but global as-rigid-as possible.
However, these methods above have limited ability to model and manipulate the complicated scene geometry for further visual effect rendering.

\begin{figure*}
	\includegraphics[width=\linewidth]{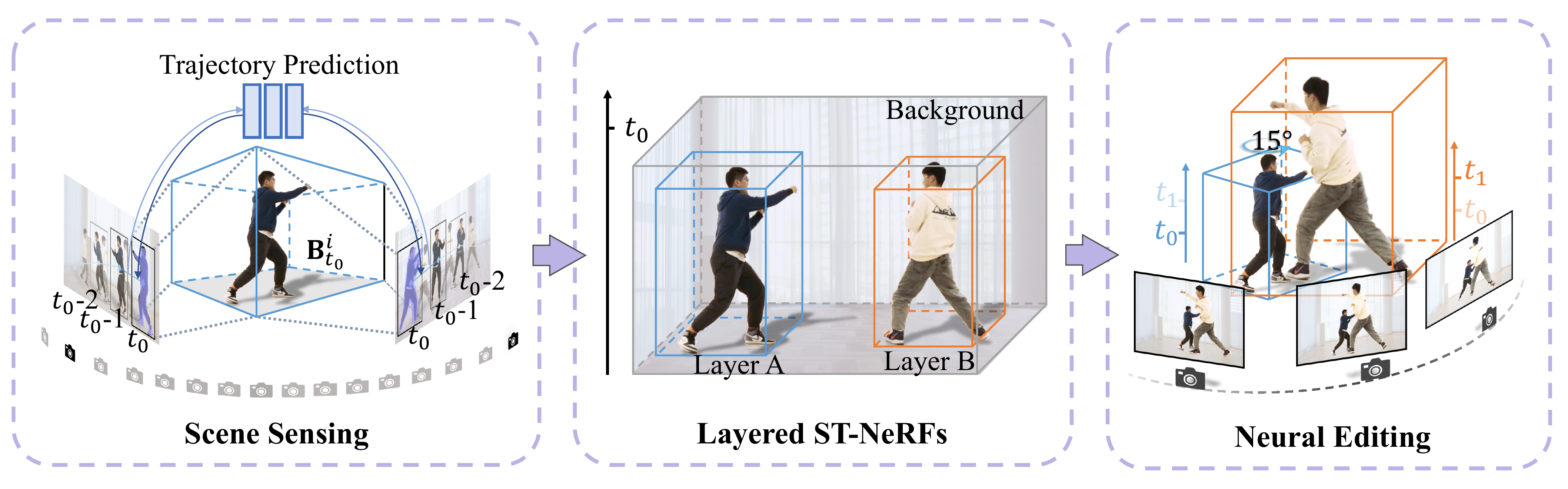}
	\caption{\textbf{The algorithm pipeline of our approach for editable free-viewpoint video generation.}
		Given the 16 synchronized RGB video covering a view range up to 180 degrees as input, our approach first includes a scene sensing stage to provide a layer-wised label map tracking with a 3D corresponding bounding box $\textbf{B}^i_{t_0}$ and initial geometry for the $i$-th dynamic entity in the captured scene at time $t_0$. 
		Then, a new layered neural representation is adopted to model each dynamic entity into a spatio-temporal coherent neural radiance field (ST-NeRF).
		Finally, various neural editing functions are introduced based on our layered representation to fully support perception and realistic manipulation of the dynamic scene.
		Our approach for the first time enables various spatial editing like affine transform or duplication as well as temporal editing like retiming performers' movements in free-viewpoint videos.}
	\Description{Fully described in the text.}
	\label{fig:pipeline}
\end{figure*}

\hspace*{\fill} \\
\noindent{\bf Neural Rendering.}
The recent progress of neural rendering techniques brings huge potential for photo-realistic novel view blending~\cite{NRWird_CVPR19,DeepBlending,FreeViewSynthesis,zhang2021neural} and constructing neural scene representations~\cite{DeepVoxels_CVPR2019,NeuralVolumes,SRN_nips2019,mildenhall2020nerf, flynn2019deepview, zhou2018stereo}.
For neural blending, various methods learn the mapping of features from source images to novel target views, where the learned additional deep features are assigned on top of reconstructed meshes~\cite{DeepBlending,FreeViewSynthesis,thies2019deferred} or the depth maps~\cite{flynn2016deepstereo,xu2019deep}, while some recent works further models the view-dependent effects~\cite{Thies2020Image-guided, DVS_photo} or large scene rendering in-the-wild~\cite{meshry2019neural}.
However, these neural blending approaches above rely on static scene modeling.
Only recently, the method~\cite{yoon2020novel} enables dynamic scene rendering by performing explicit depth-based 3D warping.
However, this method suffers from limited free-viewing range and fragile dynamic motion modeling.

For reconstructing neural scenes, various data representations have been explored, such as point-clouds~\cite{aliev2019neural,Wu_2020_CVPR}, voxels~\cite{DeepVoxels_CVPR2019,NeuralVolumes} or implicit representations~\cite{SRN_nips2019,mildenhall2020nerf,suo2021neuralhumanfvv}.
Researchers~\cite{RTH_nvs,DodgeABullet_ECCV2018} also utilize the underlying latent geometry for novel view
synthesis of human performance in the encoder-decoder manner, which still suffers from limited representation ability of a single latent code for complex human inferior texture output.
The most notable approach Neural Radiance Field (NeRF)~\cite{mildenhall2020nerf} combines the implicit representation with volumetric rendering 
to achieve compelling novel view synthesis with rich view-dependent effects.
However, these neural representations above can only handle static scenes, and the literature on dynamic scene neural representation remains sparse.
Recent work~\cite{park2020deformable,pumarola2020d,li2020neural,xian2020space,tretschk2020non,rebain2020derf,ost2020neural} extend the approach NeRF~\cite{mildenhall2020nerf} using neural radiance field into the dynamic setting. 
They decompose the task into learning a spatial mapping from the canonical scene to the current scene at each time step and regressing the canonical radiance field. 
However, the above solutions using dynamic neural radiance fields still suffer from limited capture volume or fragile human motions,
without additional editing functions for the visual effects that can change the perception of the dynamic scenes.
Recent work~\cite{lu2020layered} uses a monocular video as input, and it enables the neural retiming effects of humans. However, without multiview input, as we have, it has no free-viewing ability for space-time visual effects.
In contrast, our novel layered neural representation extends the neural radiance field to model large-scale dynamic scenes and provides unique spatial and temporal editing functions for photo-realistic visual effects while still supporting free viewing.

\hspace*{\fill} \\
\noindent{\bf Dynamic Scene Reconstruction.}
Different than the reconstruction of static scenes, tackling dynamic scenes requires settling the illumination changes and moving objects. 
To obtain a reconstruction for dynamic objects with input data from either camera array or a single camera, methods involving silhouette~\cite{taneja2010modeling, kim2010dynamic}, stereo~\cite{luo2020consistent, li2019learning, lv2018learning,FlyCapLan}, segmentation~\cite{russell2014video, ranftl2016dense}, and photometric~\cite{ahmed2008robust,vlasic2009dynamic,he2021challencap} have been explored. 
Early solutions~\cite{Mustafa_2016_CVPR,collet2015high,motion2fusion} rely on multi-view dome-based setup for high-fidelity reconstruction and texture rendering of human activities in novel views.
Recently, volumetric approaches with RGB-D sensors and modern GPUs have enabled real-time dynamic scene reconstruction and eliminated the reliance on a pre-scanned template model.
The high-end solutions~\cite{dou-siggraph2016,motion2fusion,TotalCapture,UnstructureLan} rely on multi-view studio setup to achieve high-fidelity reconstruction and rendering, while the low-end approaches~\cite{newcombe2015dynamicfusion,FlyFusion,robustfusion} adopt the most handy monocular setup with a temporal fusion pipeline~\cite{KinectFusion} but suffer from inherent self-occlusion constraint.
However, all these volumetric reconstruction suffers from the inherent limited captured volume constraint, especially for reconstructing a large-scale dynamic scene.
Differently, our approach models the dynamic scene into a layered neural representation, where all the dynamic entities, including the environment itself, are represented as several spatio-temporal coherent neural radiance fields.

\hspace*{\fill} \\
{\bf Video Editing.}
Video Editing encodes various visual effects that can change the perception of the recorded videos.
Effects like time warping and object editing can be achieved by manipulating videos. 
For instance, the representative work~\cite{goldman2008video} tracks 2D object motion to enable novel interactions with video, including annotations, navigation, or direct manipulation, creating an image from multiple video frames. 
In this paper, we focus on human-related elements editing in videos.
For human motion manipulating, various methods have been proposed for transferring motion between different people in different videos~\cite{aberman2018deep,chan2019everybody}, or manipulating the appearance from a low-dimensional motion signal like skeletons~\cite{gafni2019vid2game,liu2019neural}.
Abe~~\emph{et al.}~\cite{10.1145/3197517.3201371} manipulate the dancing appearance of performers in the video by warping and aligning the visual beats with the musical beats.
The recent work~\cite{lu2020layered} utilizes a learning-based layered video representation to manipulate the timing of the motions of different performers in the video.
The layers in \cite{lu2020layered} can be re-assembled into a new video with various retiming visual effects.
By combining such layered representation with neural rendering, high-quality rendering with temporally visual effects can be achieved. 
However, the video manipulation methods above only consider the monocular capture setting or limited in the temporal effect only, without exploring the rich 3D spatial editing functions for novel view synthesis in free-viewpoint videos.

\section{OVERVIEW}
The presented approach marries the free-viewpoint videos with a new neural layered space-time representation, which generates editable free-viewpoint videos for large-scale dynamic scenes with multiple performers. 
Our system takes only 16 synchronized RGB videos to cover a view range up to 180 degrees as input and enables various spatial and temporal editing functions for numerous fancy visual effects whilst still maintaining high realism and supporting wide-range free viewing.
Fig.~\ref{fig:pipeline} illustrates the high-level components of our approach, which models the space-time correlations of all the dynamic entities, including the environment itself, into a consistent neural representation so as to support fully perception and realistic manipulation of the dynamic scene.

\hspace*{\fill} \\
\noindent{\bf Scene Parsing.} 
We first adopt a scene parsing stage to generate the coarse space-time 4D label maps of all the dynamic performers in the captured scene from the input multi-viewpoint videos.
To this end, we utilize the inherent geometry before our multi-view setting via the patch-based Multi-view stereo (MVS) technique~\cite{luo2019p} to generate coarse dynamic point clouds for all the frames.
Then, we adopt a 4D label map tracking to generate the 3D bounding-box of each dynamic entity with moving point-clouds in each timestamp.
Such tracking combines the SiamMask~\cite{wang2019fast} tracker with the trajectory prediction network from \cite{wu2020visual} for robust position correction.
We further perform label map refinement to handle the occlusion between various performers in the camera views so as to provide the initially tracked bounding-box with coarse geometry inside as well as the layer-wise label map for each dynamic entity in each timestamp. 
\\ \hspace*{\fill} \\
\noindent{\bf Layered Neural Representation.}
The core of our approach is a novel layered neural representation for space-time photo-realistic manipulation of the dynamic scene based on the scene parsing results. 
In our neural representation, we formulate each dynamic entity as well as the environment itself into the individual neural layer to support per-entity editing.
Then, such a neural layer/entity is formulated into a continuous function of both space and time, forming a spatially and temporally consistent neural radiance field using multi-layer perceptrons (MLPs) called ST-NeRF.
In the ST-NeRF for a dynamic entity, a continuous deform module is introduced to encoded the temporal motion information across all the views and timestamps.
Besides, the tracked bounding-box of the entity serves as the inherent spatial anchor in our ST-NeRF.
Thus, we extend the ray sampling strategy in the original NeRF~\cite{mildenhall2020nerf} to a multi-segment version and propose an object-aware volume rendering scheme to re-assemble all the neural layers into the editable synthesis results in novel views. 
A motion-aware ray sampling strategy is further adapted to enable efficient training for a large dynamic scene with multiple dynamic performers.
\\ \hspace*{\fill} \\
\noindent{\bf Neural Editing.}
Our various space-time editing functions are enabled by the disentanglement of location, deformation as well as the appearance of all the dynamic entities in our layer neural representations.
Recall that the input of each ST-NeRF is the position and timing of the corresponding dynamic entity, which can be explicitly manipulated by the users during the inference.
Such basic operations of neural editing enable a series of space-time visual editing effects, while our layer-wised representation with neural radiance field enables the generation of photo-realistic free-viewpoint results.
To this end, various spatial editing like affine transform or duplication as well as temporal editing like retiming performers' movements can be applied to each layer/entity in a depth-aware and photo-realistic manner.

\begin{figure}
	\includegraphics[width=\linewidth]{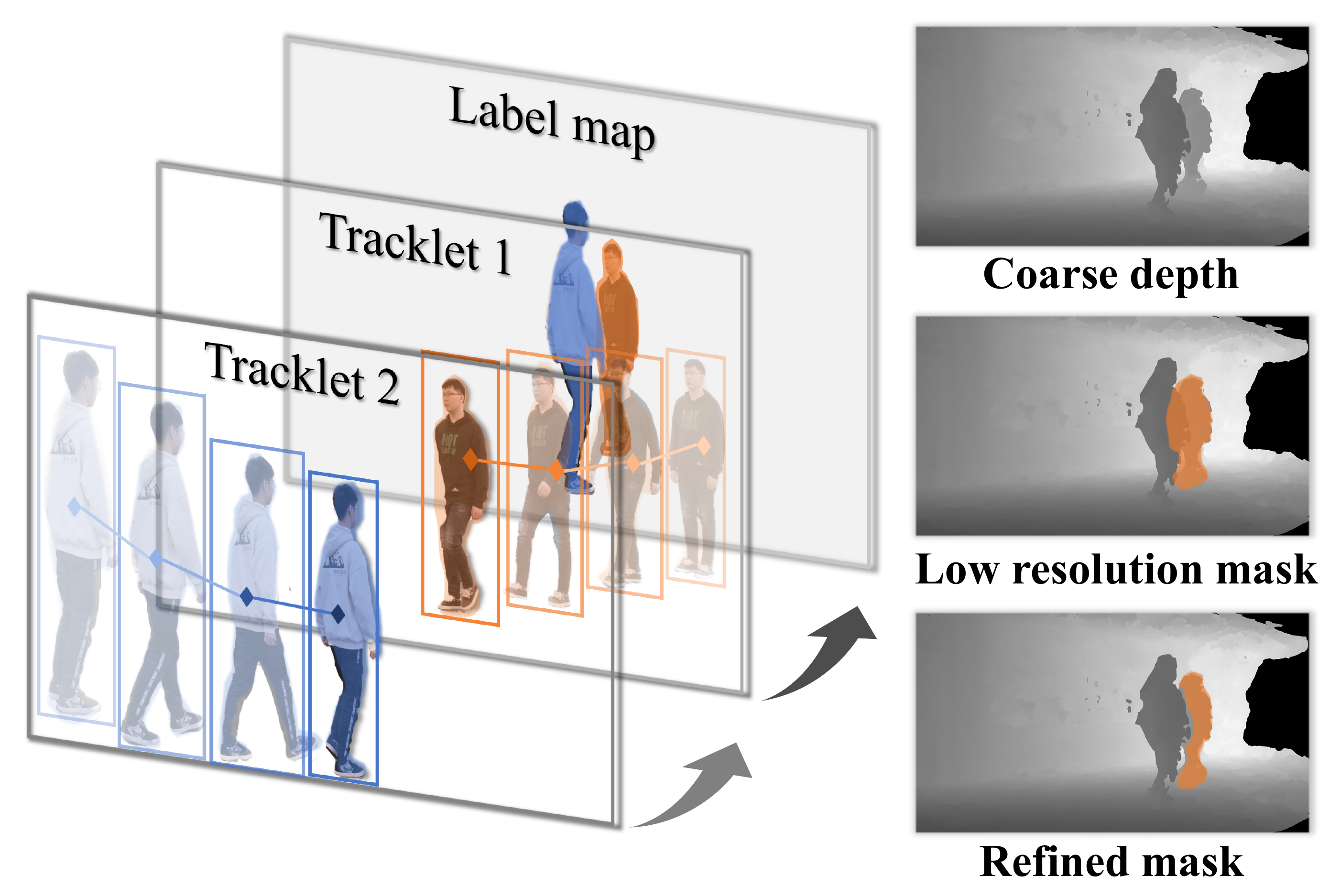}
	\caption{\textbf{Illustration of our scene parsing stage.} It provides the tracked bounding-box with coarse geometry inside as well as the layer-wise label map for each dynamic entities in each timestamp.}
	\Description{Fully described in the text.}
	\label{fig:module1}
\end{figure}

\section{METHOD}\label{sec:method}

\subsection{Scene Parsing}\label{sec:layermap}
The goal of our scene parsing stage is to generate the coarse space-time 4D label maps of all the dynamic performers in the captured scene.
Here, the label map is a per-pixel segmentation mask with human object identities in the image, denoted as $\{\mathbf{L}_t^c\}_{t=0}^{n_t}$ in camera view $c$ for a sequence with $n_t$ frames. 
A label map $\mathbf{L}_t^c$ is the merged pixel sets $\{ \mathcal{M}_t^{c,i}\}_{i=1}^{n_i}$ where $\mathcal{M}_t^{c,i}$ is a pixel set belonged to the $i$-th performer; $n_i$ is the total object number. 
What is more, we also obtain 3D axis-aligned bounding boxes of the dynamic entities denoted by $\{\mathbf{B}_t^i\}_{i=1}^{n_i}$ with a coarse geometry of each dynamic entity. The whole stage of scene parsing is represented in Fig.~\ref{fig:module1}.
\\ \hspace*{\fill} \\
{\bf Geometry Estimation.}
Since the $n_c$ calibrated cameras have enough overlaps to cover the dynamic scene, we utilize the inherent geometry prior to our multi-view setting and reconstruct a dense depth map sequence of each viewpoint using the patch-wise learning-based Multi-view stereo (MVS) method~\cite{luo2019p} under a low depth resolution (half of the input resolution). 
The reconstructed depth maps $\{ \mathbf{D}_t^c \}$ also serve as a cue for label map refinement later. 
\begin{figure*}[ht]
	\includegraphics[width=\linewidth]{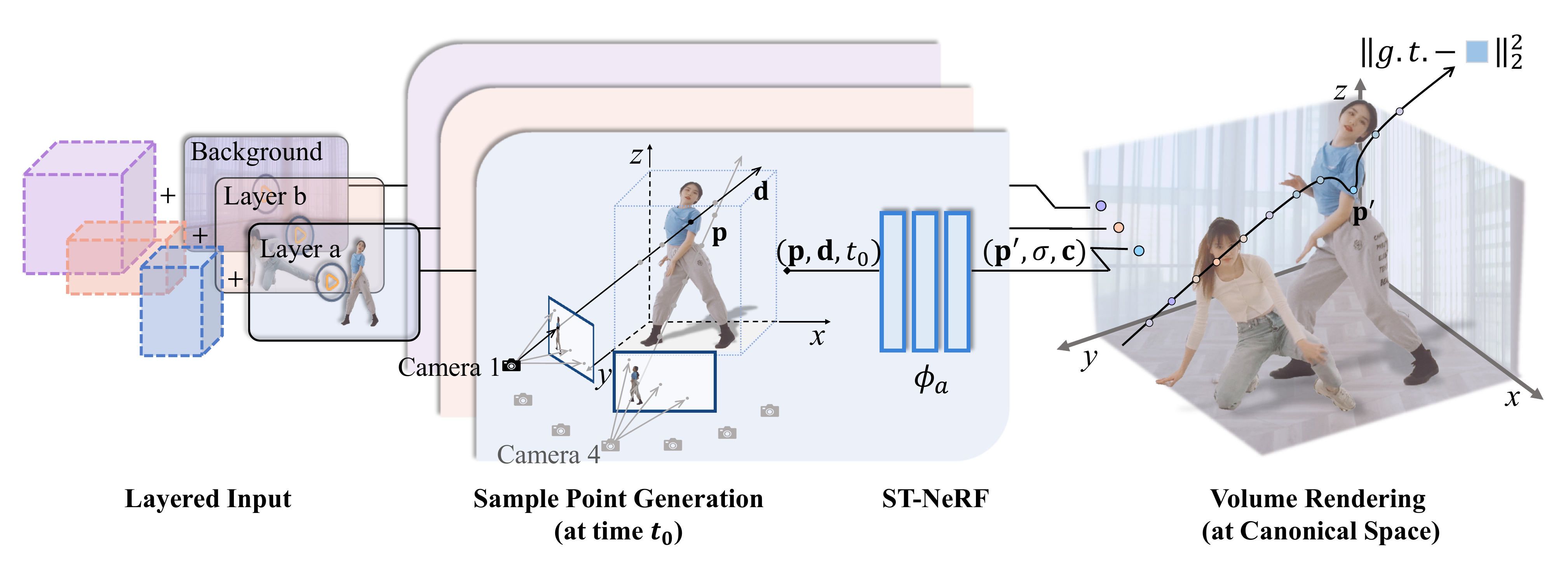} 
	\caption{\textbf{Layered Spatio-time Neural Rendering.} 
		The input consists of multi-layer videos and their bounding boxes. 
		For each layer, our approach feeds sampled points to the corresponding ST-NeRF $\phi_i$, $i\in\{a, b, bg\}$ in this example.
		Note that each sampled point has three attributes: the position $\mathbf{p}$, the direction $\mathbf{d}$, and the timestamp $t_0$.
		The output of $\phi_i$ has three attributes as well, the deformed position $\mathbf{p'}$, density $\sigma$, and color $\mathbf{c}$.
		Then, object-aware volume rendering is applied to obtain the predicted pixel color in canonical space.
		Finally, we supervise the result to be consistent with the ground truth pixel color.
	}
	\Description{Fully described in the text.}
	\label{fig:module2}
\end{figure*}
\\ \hspace*{\fill} \\
{\bf Multi-view Label Map Tracking.}
In order to track the 3D bounding boxes of a dynamic entity, we apply a multi-view visual object tracking scheme and then restore the 3D information.   
The key role of our tracking scheme is to provide human object 2D bounding boxes and trajectories, which consists of tracklets with the same identity across cameras.
However, occlusions will cause tracking failures in a multi-object scene during cross-view identity association.
To tackle this issue, we adopt the SiamMask~\cite{wang2019fast} tracker, which is a one-shot single object tracker with mask estimation to multi-view tracking.
Specifically, we manually annotate the initial 2D bounding boxes $\{\mathbf{b}_0^{c,i}\}_{c=1}^{n_c}$ for the $i$-th human objects in all the views. 
The tracker is conducted on each object separately in each view and forms tracklets $\mathcal{G}_{t_1,t_2}^c = \{{\bf g}_t^c\in\mathbb{R}^2\}_{t=t_1}^{t_2}$, where ${\bf g}_t^c$ is the center of ${\bf b}_t^c$ in a consecutive time period from time $t_1$ to time $t_2$.  
To handle occlusion scenarios and enhance multi-view constraints, we exploit a trajectory prediction network (TPN) from \cite{wu2020visual} to regularize the predicted position of the object among 2D views.
The tracking result is corrected as follows:
\begin{equation} \label{equ_trajection}
\begin{split}
 {\bf g}_t^{b'} = q^b {\bf g}_t^b + \frac{1-q^b}{w} \hspace{-2mm}\sum\limits_{c, q^c\geq\tau}\hspace{-1mm} q^c \Theta_\mathrm{TP}(\mathcal{G}_{t_0,t}^c,\mathcal{G}_{t0,t1}^b),
\end{split}
\end{equation}
where $w = \sum_{c, q^c\geq\tau} q^c$ is a normalized coefficient and $\Theta_\mathrm{TP}(\cdot, \cdot)$ is the TPN for predicting the object location ${\bf g}_t^b$ in camera $b$. 
Here we regard the confidence score $q^c$ from the last softmax layer of the tracker's result as the criterion to trigger tracking correction.
Since we track each human object independently, tracklets will keep identity consistent and associated with camera views, while TPN also manages to avoid identity-switch by exploiting multi-view data.  
With the above robust tracking results, we further estimate the 3D bounding boxes of each performer in the scene. 
Once we have 2D bounding boxes of each object, we use them as the silhouette masks to reconstruct coarse geometry of individual human using the shape from space carving algorithm~\cite{kutulakos2000theory}. 
Then, we associate the desired 3D bounding box of a human object with this coarse geometry tightly.
\\ \hspace*{\fill} \\
{\bf Label Map Refinement.}
The tracker also predicts the mask of the human object in all the views during tracking. 
However, this kind of predicted masks is rather rough due to foreground occlusion.
Specifically, the occluded pixels will be associated with various performers in the mask, disturbing our layered neural rendering.  
To this end, we refine human masks using a statistical method. 
Assuming that we have the refined mask of the target object in the previous frame, the averaged depth value of the target person can be calculated according to the reconstructed depth map $\{ \mathbf{D}_t^c \}$. 
Then, for the mask from the current frame, we discard the pixels whose depth value is deviated from the averaged depth in the previous frame inside the current mask to obtain a refined mask.  
All these refined masks are further composited into label maps $\{ \mathcal{M}_t^{c,i}\}$. 
Especially, we also regard the background scene as a special object whose 3D bounding box can be calculated from the entire scene 3D point cloud.

\subsection{Spatio-Temporal neural radiance field}

The core of our approach is a new layered neural representation where each tracked entity is formulated as a separated continuous function of both space and time, forming a spatially and temporally consistent neural radiance field (ST-NeRF) to support photo-realistic editing functions. The module is illustrated in Fig.~\ref{fig:module2}.
Recall that the original neural radiance field (NeRF)~\cite{mildenhall2020nerf} is a continuous representation for mapping each 3D point $\mathbf{p} = (x,y,z)$ and a viewing direction $\mathbf{d}=(d_1,d_2)$ to the density $\sigma$ and the color $\mathbf{c}=(r,g,b)$. 
Differently, our ST-NeRF models the space-time coherence between the dynamic entity and the scene, which implicitly records the motion, geometry, and appearance information of the performer based on the corresponding tracking results from the previous stage. 

To this end, our ST-NeRF is parameterized as MLP networks $\phi$, which consists of two modules: a space-time deform module $\phi^d$ and a neural radiance module $\phi^r$. 
$\phi^d$ deforms sample points from various time and space into a canonical space, while $\phi^r$ records the geometry and color of the dynamic entity.
Similar to \cite{pumarola2020d}, we adopt a MLP-based deformation network for $\phi^d$ to handle dynamic scenes. 
Instead of using latent codes to encode frames, we adopt a more efficient way, where the frame number is directly encoded into a high dimension feature without any computing and storage overhead by using positional encoding~\cite{mildenhall2020nerf}. 
Specifically, $\phi^d$ and $\phi^r$ in ST-NeRF cooperate in the following way:
\begin{equation}
\begin{split}
\phi^{r}(\mathbf{p}+\Delta\mathbf{p}, \mathbf{d}, t, \theta^r) &= (\mathbf{c}, \sigma) \\
\Delta\mathbf{p} &= \phi^{d}(\mathbf{p}, t, \theta^d),
\end{split}
\label{eq:ST-NeRF}
\end{equation}
where $\theta^r$ and $\theta^d$ are network weights. Notably, all inputs except network weights are mapped by a positional encoding function.

\begin{figure}[t]
	\includegraphics[width=\linewidth]{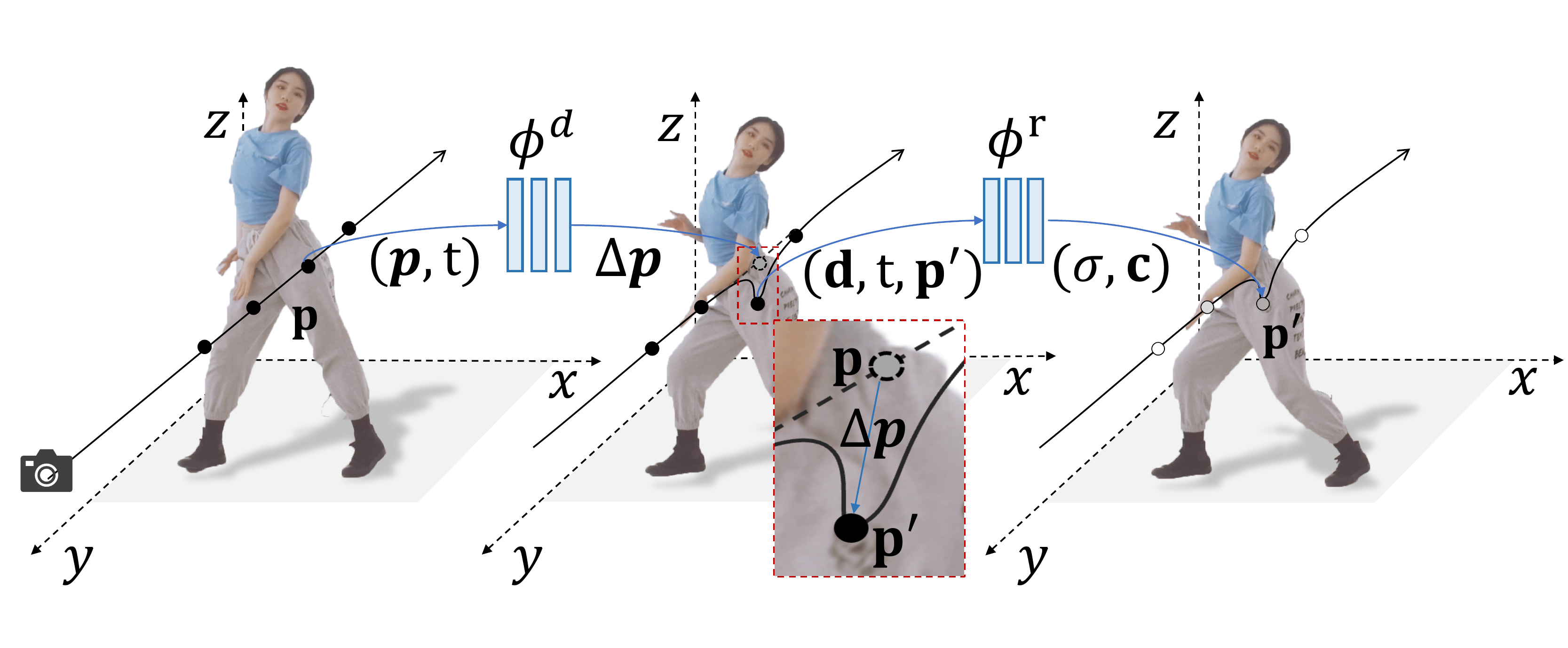}
	\caption{\textbf{ST-NeRF.} The figure illustrates the procedure of obtaining attributes using our model, including a deformed position in canonical space, density and color. $\phi^d$ first deforms the position from original space to canonical space. Feeding the deformed position $\mathbf{p'} = \mathbf{p} + \mathbf{\Delta p}$, original ray direction $\mathbf{d}$ and timestamp $t$, $\phi^r$ yields the density $\sigma$ and the color $\mathbf{c}$ at position $\mathbf{p'}$ in radiance field.}
	\Description{Fully described in the text.}
	\label{fig:inference}
\end{figure}

%
The Neural representation of a dynamic entity is given by $\Theta=\{\theta^r, \theta^d\}$. 
Note that the neural radiance module $\phi^r$ also inputs a timestamp to handle time-dependent appearance change. 
The network pipeline of our ST-NeRF is illustrated in Fig.~\ref{fig:inference}, which is formulated as follows:
\begin{equation}
\begin{split}
\phi(\mathbf{p}, \mathbf{d}, t, \Theta) &= (\mathbf{c}, \sigma).
\end{split}
\label{eq:ST-NeRF_overall}
\end{equation}

Note that the visibility problem is a fundamental challenge for dynamic scene modeling with multiple dynamic entities, where some local appearance information of various entities will be missing due to occlusion or view foreshortening in the observed views.
In ST-NeRF, both geometry and appearance information across views and timestamps are fused in the canonical space in an effective self-supervised manner.
Thus, ST-NeRF potentially can handle the inherent visibility challenge and provide a complete and photo-realistic novel-view synthesis. 

\subsection{Layered Spatio-Temporal Neural Renderer}\label{sec:Renderer}
Recall that in our ST-NeRF for a dynamic entity, the deform module encodes motion information across all the views and timestamps, while the tracked bounding-box of the entity serves as the inherent spatial anchor.
To this end, we extend the ray sampling strategy in the original NeRF~\cite{mildenhall2020nerf} to a multi-segment version and propose an object-aware volume rendering scheme to re-assemble all neural layers.
Such a layered spatio-temporal neural renderer enables photo-realistic novel view synthesis where each neural layer/entity is fully editable.
Specifically, our rendering pipeline contains three steps as described in the following.
\begin{figure}[t]
	\includegraphics[width=\linewidth]{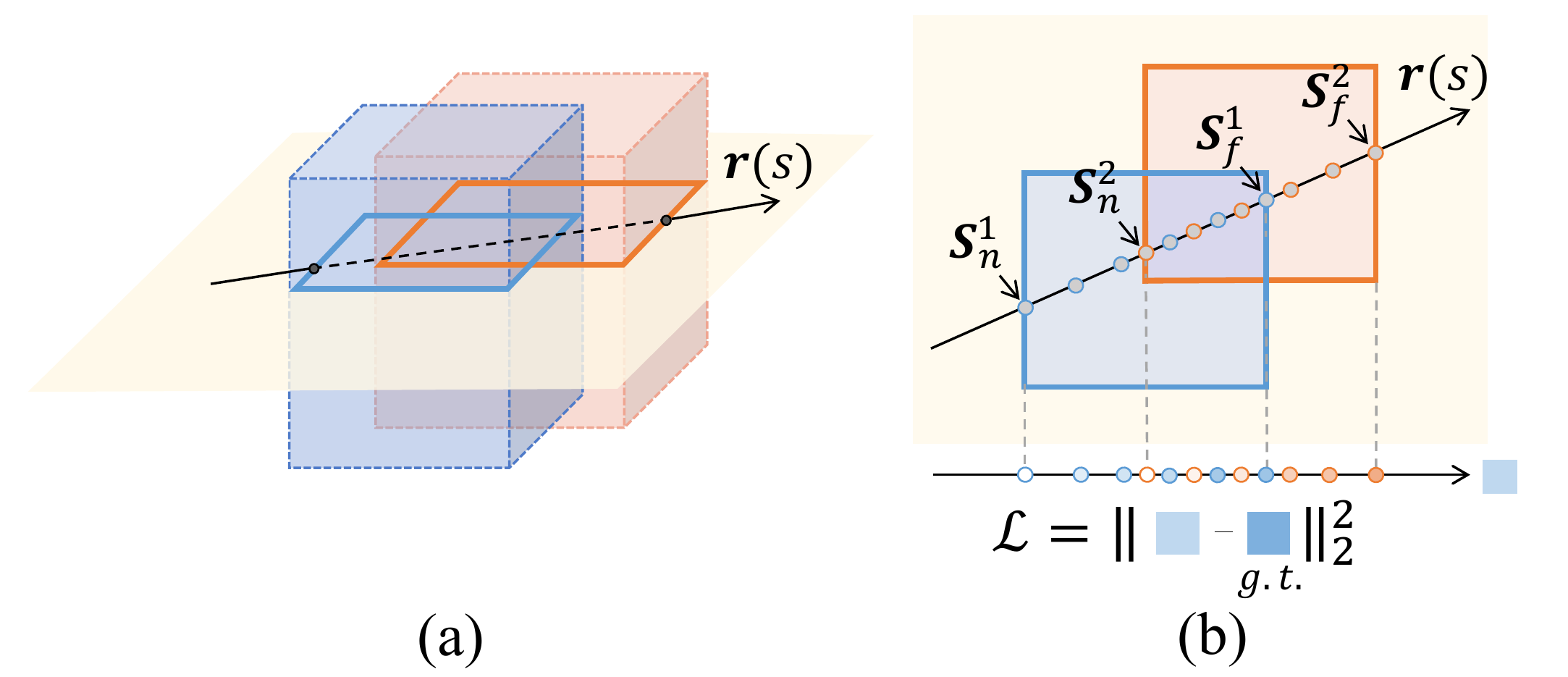}
	\caption{\textbf{Layered Ray Sampling Scheme.} For a ray passing through multiple bounding boxes, we sample points for each bounding box, feeding points to corresponding ST-NeRF to get densities and colors. We obtain the predicted pixel color with volume rendering. Finally, we calculate the L2-norm of predicted color with ground truth color as RGB loss.}
	\Description{Fully described in the text.}
	\label{fig:sampling} 
\end{figure}
\begin{figure*}[th]
  \includegraphics[width=\linewidth]{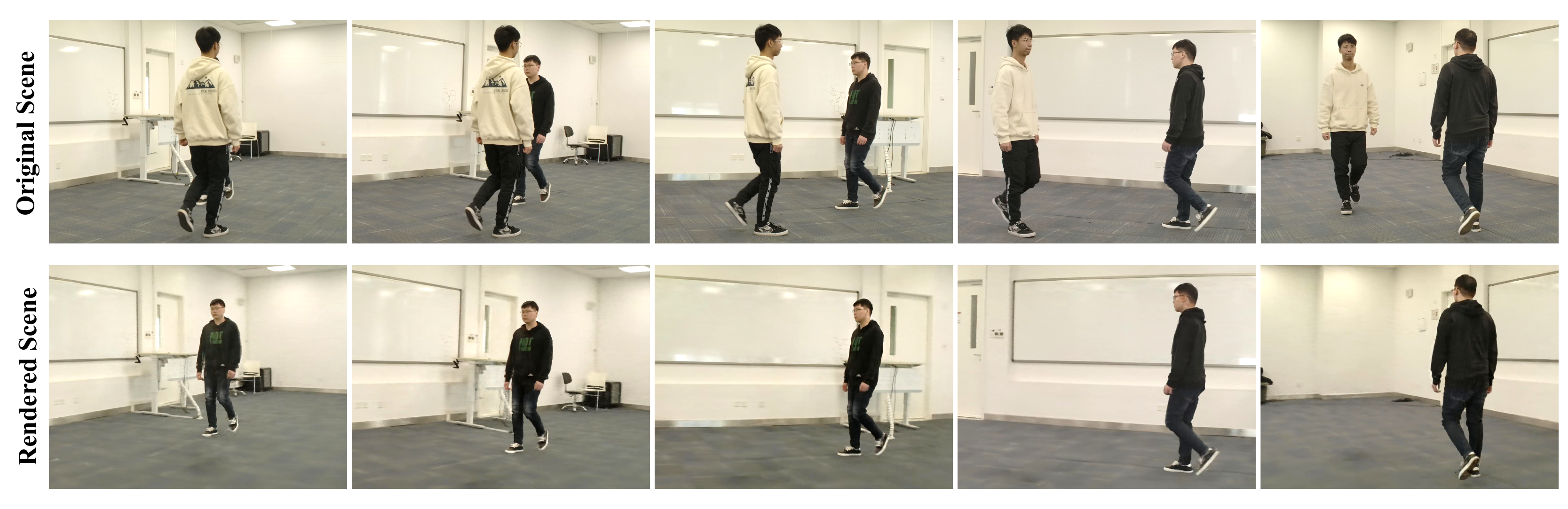}
  \caption{\textbf{Occluded Layer Rendering.} 
  The images on top are the input from five different cameras, and the bottom row shows our rendering result removing the performer in the front.
  Our approach can handle neural layer-wise occlusion and infer occluded layer appearance.
  }
  \Description{Fully described in the text.}
  \label{fig:occluded}
\end{figure*}
\\ \hspace*{\fill} \\
{\bf Scene Composition.}
Since our layered ST-NeRF representation has disentangled the explicit pose of the entity from implicit object geometry and appearance, we can further manipulate the corresponding 3D bounding boxes of various entities at any position in the scene. 
Actually, such feature deeply enables the numerous editing effects, which we will introduce later in Sec.~\ref{sec:Scene Editing}. 
At the beginning of our rendering pipeline at a timestamp $t$, we composite the target scene by determining placements of 3D bounding boxes $\mathbf{B}_t^i$ for the $i$-th dynamic entity. 
Next, we set up a virtual camera in the scene and generate camera rays passing through the edited scene.
\\ \hspace*{\fill} \\
{\bf Ray Segmentation and Sampling.}
Instead of treating an entire camera ray equally during point sampling, we divide the camera ray into object-level segments and deploy the same sampling strategy to each segment, respectively. 
To this end, for a camera ray $\mathbf{r}(s)$, we compute its intersections with each 3D bounding box $\mathbf{B}_t^i$ of the $i$-th entity at current timestamp $t$ and obtain an segment $\mathcal{S}^i=\{s_n^i, s_f^i | s_n^i < s_f^i\}$, where $s_n^i$ and $s_f^i$ are depth values of intersection points. 
Note that a segment $\mathcal{S}^i$ is valid if and only if there are two different intersections on camera ray, and
the indexes of all valid segments' objects are denoted by $\mathcal{I}$. 

Similar to the original NeRF~\cite{mildenhall2020nerf}, we deploy a hierarchical sampling strategy on every valid segment. 
In the coarse sampling stage, we partition each segment into $N$ evenly-spaced bins and draw one sample point from each bin uniformly at random:
\begin{equation}
   s_{j}^i\sim\mathcal{U}\left[s_{n}^i+\frac{j-1}{N}(s_{f}^i-s_{n}^i), s_{n}^i+\frac{j}{N}(s_{f}^i-s_{n}^i)\right], j\in\left[1, 2, ..., N\right],
\end{equation}
where $s_{j}^i$ is the depth value of $j$-th sampled point on the ray. 

Let $\mathcal{P}_c^i=\{\mathbf{r}(s_{j}^i)| j\in\left[1, 2, ..., N\right] \}$ denote the sampled points for the $i$-th entity in the coarse sampling stage. 
Instead of feeding all sampled points into the same network, we adopt various ST-NeRF to infer the attributes of these points associated with various corresponding entities. 
Next, we perform a second sampling based on the probability density distribution function calculated from the sampled points' densities in the coarse sampling stage using inverse transform sampling, where $\mathcal{P}_f^i$ denotes these sampled points in the fine stage. 
Fig.~\ref{fig:sampling} illustrates our ray segmentation and sampling strategy using a single ray as example. 
\\ \hspace*{\fill} \\
{\bf Object-aware Volume Rendering.}
Given sampled points $\mathcal{P}^i$ along the ray $\mathbf{r}$ and their predicted densities and colors, the final color $\hat{\mathbf{C}}(\textbf{r})$ of the pixel is integrated in an object-aware manner to assemble all the neural layers into photo-realistic results. 
Specifically, before integration, the sampled points from valid segments are merged to form a point set $\mathcal{P}$ which is formulated as:
\begin{equation}
   \mathcal{P} = \bigcup_{i\in\mathcal{I}}\mathcal{P}^i.
\end{equation}
Then, we sort these points in $\mathcal{P}=\{\mathbf{p}_j\}_{j=1}^{|\mathcal{P}|}$ using their depth values from near to far. 
Thus, the final color $\hat{\mathbf{C}}(\textbf{r})$ is formulated as:
\begin{equation}
\begin{split}
    \hat{\mathbf{C}}(\textbf{r})&=\sum_{j=1}^{|\mathcal{P}|} T(\mathbf{p}_j)\left[(1-\text{exp}(-\sigma(\mathbf{p}_j)\delta(\mathbf{p}_j))\mathbf{c}(\mathbf{p}_j)\right]\\
    T(\mathbf{p}_j)&=\text{exp}\left(   -\sum_{k=1}^{j-1}\sigma(\mathbf{p}_k)\delta(\mathbf{p}_k)    \right),
\end{split}
\end{equation}
where $\delta(\mathbf{p}_j)=\mathbf{p}_{j+1} - \mathbf{p}_j$ is the distance between adjacent samples.  
Particularly for the rendering following the coarse sampling stage, $\mathcal{P}^i = \mathcal{P}_c^i$. 
For hierarchical sampling and rendering, both $\mathcal{P}_c$ and $\mathcal{P}_f$ in the second stage are merged for the integration, where $\mathcal{P}^i = \mathcal{P}_c^i  \cup \mathcal{P}_f^i$. 
As shown in Fig.~\ref{fig:occluded}, our object-aware volume rendering can handle neural layer-wise occlusion.

\begin{figure*}[thb]
  \includegraphics[width=\linewidth]{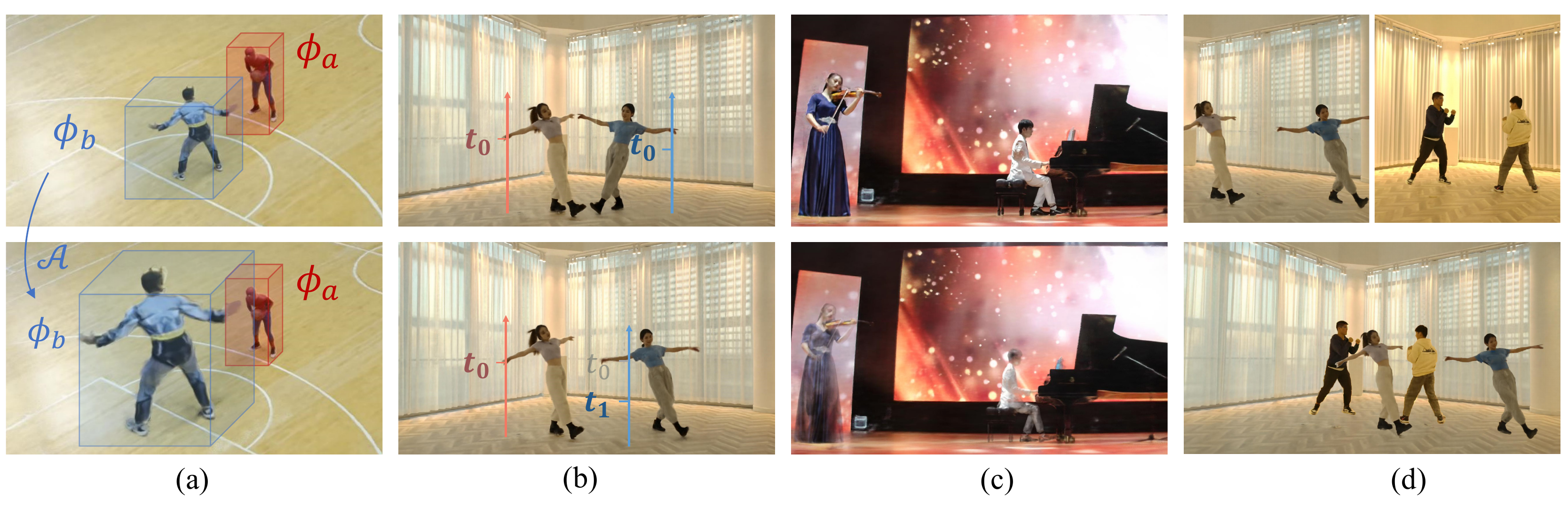}
  \caption{\textbf{Scene Editing Results.} ST-NeRF enables a variety of editing operations. \textbf{(a)} Spatial Affine Transformation, \textbf{(b)} Temporal Retiming Transformation, \textbf{(c)} Transparency Adjustment and \textbf{(d)} Object Insertion and Removal.}
  \Description{Fully described in the text.}
  \label{fig:module3}
\end{figure*}

\subsection{Network Training}\label{sec:training}
Here we introduce an effective training scheme of our layered neural representation, especially for a large dynamic scene with multiple dynamic performers.
Note that we train all the ST-NeRF networks for all the dynamic entities together so as to assemble all the neural layers and enable self-supervised training. 
To this end, since we utilize a hierarchical sampling and rendering strategy to render the scene frame by frame, the RGB loss function $\mathcal{L}_{rgb}$ is formulated as:
\begin{equation}
    \mathcal{L}_{rgb} = \sum_{\mathbf{r} \in \mR}(\|C(\mathbf{r})-\hat{C}_c(\mathbf{r})\|_2^2 + \|C(\mathbf{r})-\hat{C}_f(\mathbf{r})\|_2^2),
\end{equation}
where $\mR$ is the set of sampled rays in the mini-batch; $\mathbf{C}(\mathbf{r})$ is the ground truth color of the camera ray; $\hat{\mathbf{C}}_c(\mathbf{r})$ and $\hat{\mathbf{C}}_f(\mathbf{r}) $ are rendered colors from the coarse stage and fine stage, respectively.

As illustrated in Fig.~\ref{fig:sampling}, the object-level segments may have overlapping areas due to the intersection of 3D bounding boxes. 
Thus, sampled points from the overlapping area are utilized jointly to supervise the training of their corresponding ST-NeRF networks in our formulation, where the optimization will force the networks to determine which object these overlapping points belong to implicitly. 
In other words, our approach can handle object intersection scenarios and learn more accurate layer segmentation, enabling more photo-realistic novel view synthesis.

We also leverage the occupancy priors from the object masks $\mathcal{M}$ to accelerate network training. 
Assume that all the dynamic entities in the scene are opaque without transparency; we thus expect the layered integrated alpha values on each object's pixels to be as close to $1.0$ as possible. 
According to this insight, we design a layered loss to supervise the training of each ST-NeRF:
\begin{equation}
\begin{split}
    \mathcal{L}_{layer} &=   \frac{1}{2}\sum_{i=1}^{n_i}\|  \Omega(\mathbf{r}, \mathcal{L}, i) -\alpha(\mathbf{r}, i)\|_2^2 \\
    \alpha(\mathbf{r}, i)&=\sum_{j=1}^{|\mathcal{P}^i|}\text{exp}\left(-\sum_{k=1}^{j-1}\sigma(\mathbf{p}_k)\delta(\mathbf{p}_k)    \right)(1-\text{exp}(-\sigma(\mathbf{p}_j)\delta(\mathbf{p}_j)),
\end{split}
\end{equation}
where $\Omega(\mathbf{r}, \mathbf{L}, i)$ is a customized indicator function which outputs $1.0$ if the current pixel in the label map $\mathbf{L}$ belongs to the $i$-th entity and $0.0$ otherwise. 
Here we only use the sample points from the single ray segment of $i$-th entity to integrate alpha value $\alpha(\mathbf{r}, i)$. 

The total loss function is the linear combination of $\mathcal{L}_{rgb}$ and $\mathcal{L}_{layer}$, formulated as:
\begin{equation}
    \mathcal{L} = (1-\lambda)\mathcal{L}_{rgb} + \lambda\mathcal{L}_{layer},
\end{equation}
where $\lambda$ is the weight ratio to balance two losses, and it is dynamically adjusted during warm-up training. 
Specifically, our network training has three warm-up epochs, where $\lambda$ is set to $0.1$, $0.05$, and $0.01$ in these three epochs, respectively. 
After warm-up stage, we set $\lambda=0$. 
Such warm-up strategy provides a good initial solution for network optimization.

Note that for free-viewpoint videos with a wide viewing range, there exists a huge imbalance between the static background content and the moving dynamic content.
Training our neural representation without treating the static background and moving entities will cause training inefficiency. 
That is because, compared to objects in the scene, a large number of background camera rays lead to an imbalance in observed views and plenty of training time. 
Thus, a motion-aware strategy is further adopted to improve our training efficiency.
Specifically, we calculate proportions of each object's pixels in the dataset according to the label maps $\mathbf{L}$ obtained from scene parsing. 
Then, we resample camera rays according to their proportions with a regulation that ensures that the number of background camera rays is relatively equal to those of non-background contents. 
This simple yet efficient scheme makes the training process much easier to converge with performance improvement.

Note that our approach requires per-scene training to obtain the layered neural representation from 16 synchronized RGB videos and corresponding camera calibration data. 
We optimize our models using Adam optimizer with a learning rate that decays from $1e-4$ to $1e-5$ gradually during training.
Besides, we sample 3000 camera rays for each mini-batch which are picked up from different views and time instances randomly. 

\section{Neural Scene Editing}\label{sec:Scene Editing}
\begin{figure}[t]
  \includegraphics[width=\linewidth]{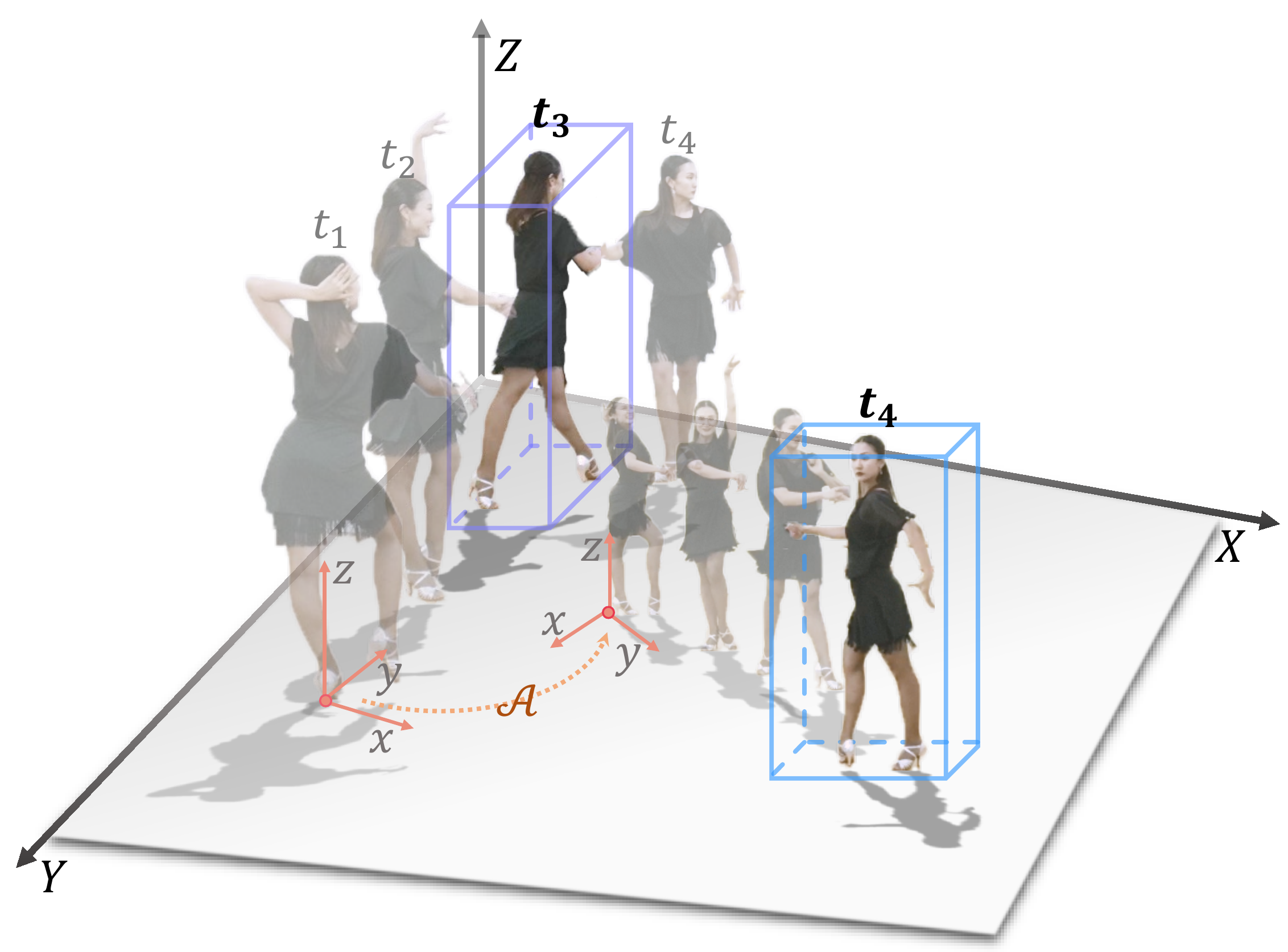}
  \caption{\textbf{Neural Scene Editing.} To enable spatio-temporal editing for various entities, we apply transformations on both timestamp and 3D bounding box of the target object.}
  \Description{Fully described in the text.}
  \label{fig:editing} 
\end{figure}

Once our layered neural representation is trained for a dynamic scene, during the inference, our layer-wise design and the disentanglement of location, timing, deformation as well as the appearance of all the dynamic entities enables fully controllable space-time visual effect editing. Our neural representation further ensures the high realism and wide viewing range of such neural editing, enabling impressive editable free-viewpoint video generation (see Fig.~\ref{fig:module3}, Fig.~\ref{fig:editing}).

To this end, our neural scene editing includes the basic operation of manipulating the input position and timing of the ST-NeRF representation of each dynamic entity.  
Such explicit space-time disentanglement enables a variety of editing functions upon each dynamic entity, such as spatial editing like affine transform or duplication as well as temporal editing like retiming performers' movements in a depth-aware and photo-realistic manner.
More importantly, all these editing effects can be produced by the combination of basic operations on various neural layers without additional training or processing.
\\ \hspace*{\fill} \\
{\bf Spatial Affine Transformation.} 
Recall that the tracked bounding-box serves as a spatial anchor to fuse the motion and appearance information of the dynamic entity across views and timestamps.
Thus, applying various affine transformations to the 3D bounding boxes is equal to re-arrange the neural canonical spaces of various entities into the current view space. 
Further combined with our layered spatio-temporal neural renderer in Sec.~\ref{sec:Renderer}, various photo-realistic spatial editing functions can be achieved, such as re-arranging the locations or scales of the individual entity in the scene. 
Note that such spatial affine transformation involves the scene composition and object-aware volume rendering processes. 
Specifically, given an affine transformation $\mathcal{A}$, we first apply it on the 3D bounding box of the $i$-th target performer and obtain a new bounding box $\hat{\mathbf{B}}^i = \mathcal{A} \circ \mathbf{B}^i$. 
Then, such $\hat{\mathbf{B}}^i$ is put into the scene to replace the original one via the same object-aware volume rendering.
After ray sampling, we apply an inverse transformation on both the sample points from the $i$-th target performer and the view direction vector $\mathbf{d}$ before feeding them into ST-NeRF, which is formulated as:
\begin{equation}
\begin{split}
\phi(\mathcal{A}^{-1}\circ\mathbf{p},\; \mathcal{A}^{-1}\circ\mathbf{d},  t, \Theta^i) &= (\mathbf{c}, \sigma),
\end{split}
\label{eq:ST-NeRF_affine}
\end{equation}
where $\mathbf{p}$ comes from the internal region of the new 3D bounding box $\hat{\mathbf{B}}^i$.
\\ \hspace*{\fill} \\
{\bf Temporal Retiming Transformation.} 
Similar to spatial editing, temporal editing functions can change the timeline of various dynamic entities so as to change the visual perception of the whole free-viewpoint video.
More specifically, the retiming transformation $\mathcal{T}$ transforms a timestamp to another discrete timestamp. 
Then, the same processes for scene composition and object-aware volume rendering are also involved in this transformation. 
During the scene composition process, we utilize the 3D bounding box at the retiming timestamp $\mathcal{T}\circ t$ for the $i$-th dynamic target, namely $\mathbf{B}_{\mathcal{T}\circ t}^i$. 
We also transform the timestamp when we inference densities and colors of sample points from object $i$ as following:
\begin{equation}
\begin{split}
\phi(\mathbf{p}, \mathbf{d},\; \mathcal{T}\circ t, \Theta^i) &= (\mathbf{c}, \sigma).
\end{split}
\label{eq:ST-NeRF_retiming}
\end{equation}
\hspace*{\fill} \\
{\bf Object Insertion and Removal.}
Note that the object insertion or removal operations in our neural editing framework only involve the scene composition process. 
To this end, we insert the new 3D bounding boxes of target performers into the scene coordinates or remove the existing ones during the scene composition process.
Our volume rendering process makes sure that such editing results maintain high realism. 
\\ \hspace*{\fill} \\
{\bf Transparency Adjustment.} 
Note that the utilized neural radiance field inherently models the transparency of individual dynamic entities.
Such transparency can also be editing naturally in our neural editing framework without any extra training, enabling more fancy visual effects.
After we obtain densities of sample points from the target object using Eqn.~\ref{eq:ST-NeRF_overall}, we can achieve the translucent effect by scaling the density value with a scalar $s$, where new density is given by $\sigma' = s \cdot \sigma$. 
We defined some simple yet effective editing functions using combinations of these basic operations. 
\begin{figure}[t]
	\includegraphics[width=\linewidth]{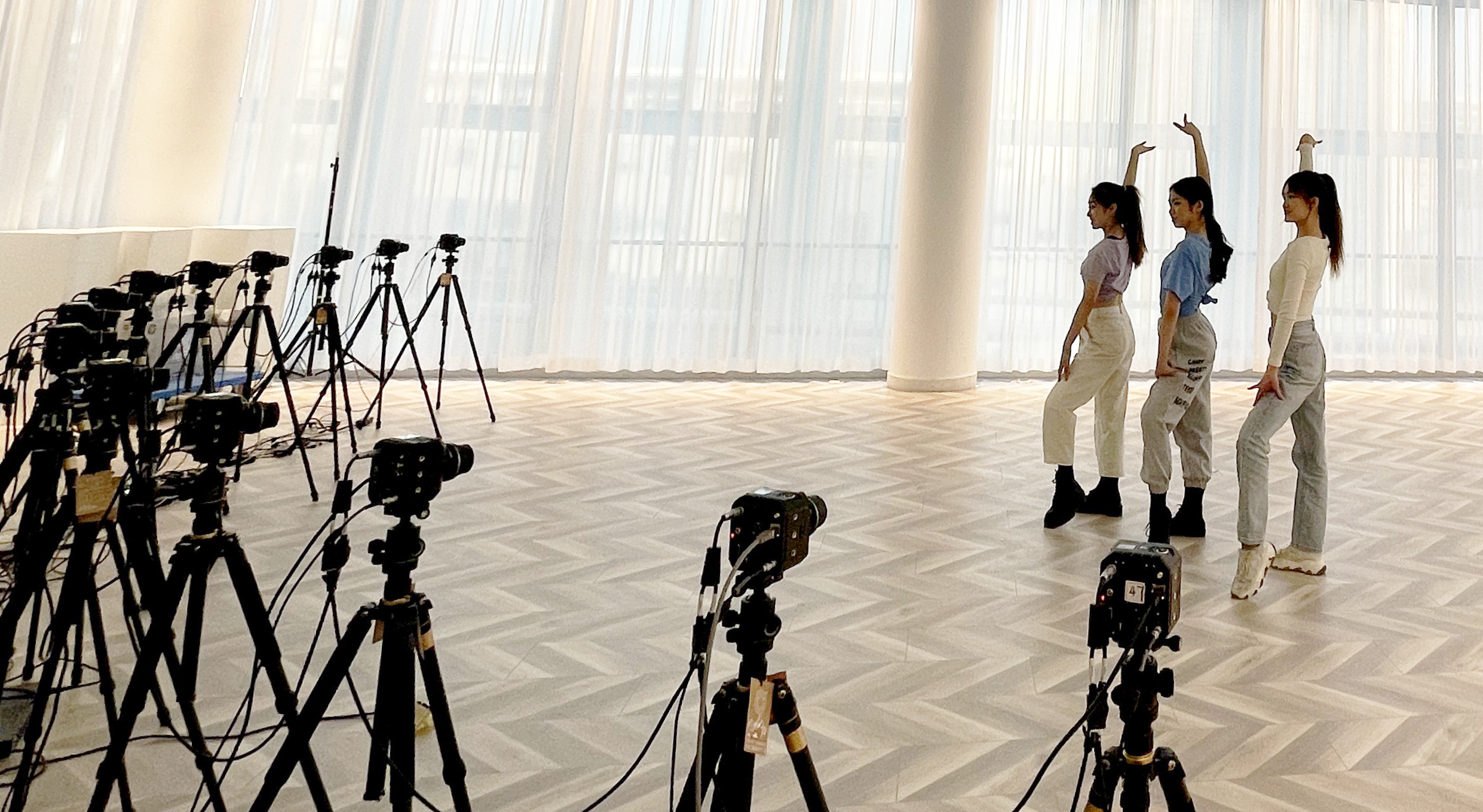}
	\caption{\textbf{Illustration of our capture system.} It consists of 16 RGB cameras to cover a view range up to 180 degrees for various dynamic scenes. All the cameras are synchronized and fixed during capturing.}
	\Description{Fully described in the text.}
	\label{fig:setting} 
\end{figure}
\section{RESULTS}\label{sec:results}
In this section, we evaluate our approach in a variety of challenging scenarios. 
We first report the implementation details of our approach and the utilized dataset captured by our multi-view system, followed by analyzing our results with various editing effects.
We further provide the comparison with previous state-of-the-art methods and the evaluation of our main technical components, both qualitatively and quantitatively. 
The limitation and discussions regarding our approach are provided in the last subsection.

\hspace*{\fill} \\
\noindent{\bf Implementation Details.}
Our network model is implemented in PyTorch.
We run all of our experiments with a single NVidia GeForce RTX3090 GPU. 
Depending on the number of video frames and neural layers in the captured scene, the training time ranges from 12 to 36 hours, with 960 $\times$ 540 cropped input image resolution. Then, we can refine the network by training it on 1920 $\times$ 1080 videos for one or two epochs. It usually takes an extra two or three days. Such a training scheme can help us save training time. Additionally, rendering a 1920 $\times$ 1080 image with three layers takes around 2 minutes.
\\ \hspace*{\fill} \\
\noindent{\bf Dataset.} 
To evaluate our method, we capture a new multi-view dataset with eight large-scale indoor dynamic scenes with two or three performers.
As shown in Fig.~\ref{fig:setting}, our capture system consists of 16 industrial Z-CAM cameras, which are uniformly arranged around a semicircle roughly towards the performers to cover a view range up to 180 degrees.
All the cameras are calibrated and synchronized in advance, producing 16 RGB streams at 1920 $\times$ 1080 resolution and 25 frames per-second.
The numbers of frames range from 75 to 350 to include several challenging human motions.
For further quantitative evaluation against our variations, we generate a synthetic system with the ground truth of label maps, bounding boxes, and camera parameters.
Specifically, we set 36 virtual cameras and to support a view range up to 360 degrees. 
Our synthesized sequence includes two dancing virtual characters with large motions lasting for 4 seconds at 25 fps.

\begin{figure*}[thp]
	\includegraphics[width=0.9\linewidth]{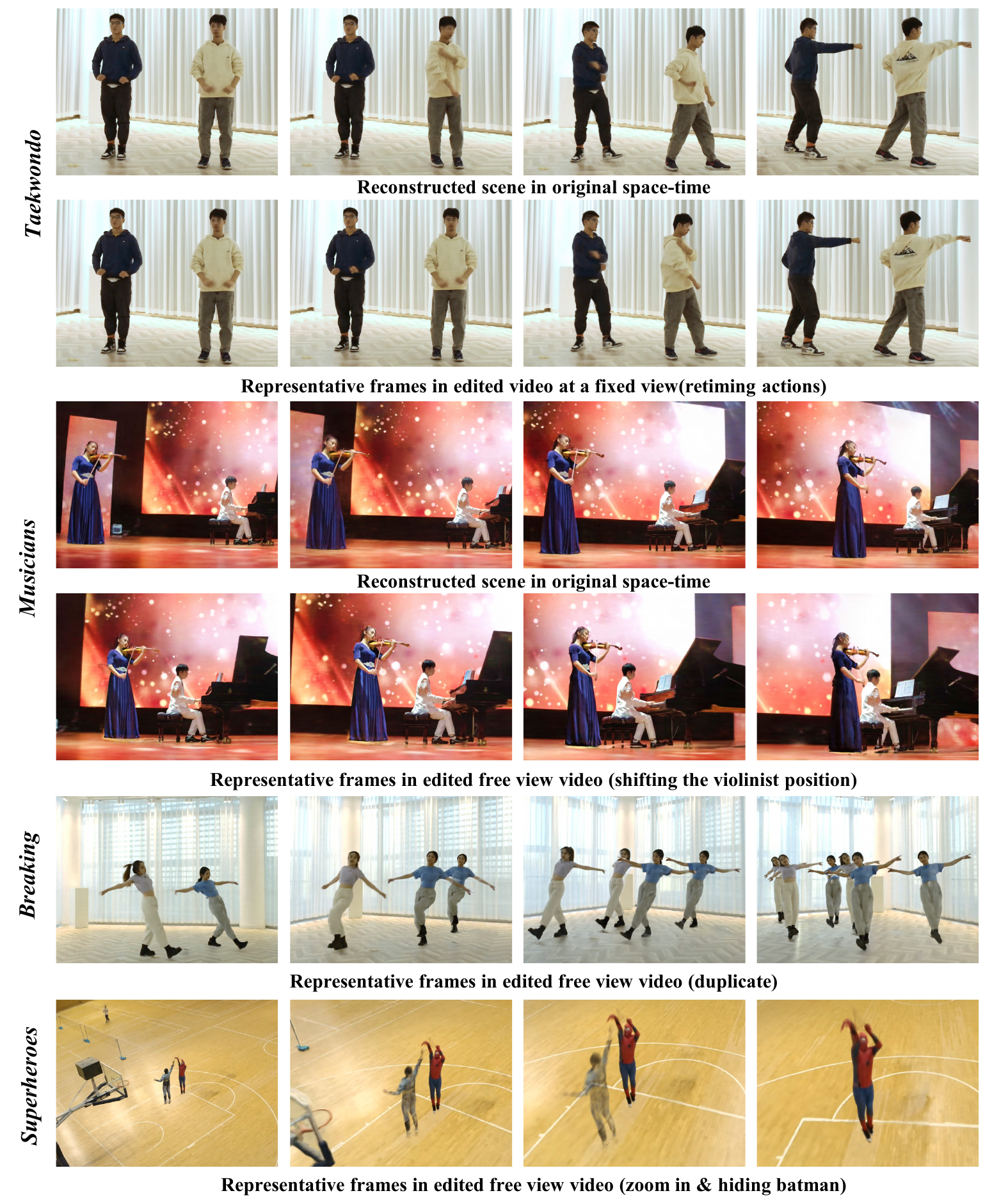}
	\caption{\textbf{Editing results.} For \textit{Taekwondo} (top), we show representative frames of fixed-viewpoint video in the original timeline and ones with retiming effects. For \textit{Musicians} (the second), we show representative frames of free-viewpoint video in the original scene, and we move the violinist to make her closer to the pianist, enabling a harmonious layout. For \textit{Breaking} (the third), we duplicate and shift dancers. Finally, for \textit{Superheroes} (bottom), we freeze time and then zoom in to the front of the spiderman with a smooth fading effect of the batman at the same time.}
	\Description{Fully described in the text.}
	\label{fig:gallery}
\end{figure*}

\begin{table}[ht]
	\begin{tabular}{@{}lllllllllllll@{}}
		\toprule
		\multicolumn{5}{c}{Comparison between different methods}                                         \\ \midrule
		\multicolumn{2}{l}{Method}                              & PSNR($\uparrow$) & SSIM ($\uparrow$ ) & MAE ($\downarrow$) & LPIPS ($\downarrow$)\\
		\multicolumn{2}{l}{NeRF}                                    & {21.7952} & {0.8755} & {0.0574} & {0.2961}\\
		\multicolumn{2}{l}{NeRF-T} 
		& {28.2553} & {0.9243} & {0.0219} & {0.2560}\\
		\multicolumn{2}{l}{Neural Volumes}                          & {28.0850} & {0.9110} & {0.0243} & {0.2608}\\
		\multicolumn{2}{l}{AGI} 
		& {14.8220} & {0.8764} & {0.0839} & {0.4543}\\
		\multicolumn{2}{l}{HVR} 
		& 24.0342 & 0.9113 & 0.0247 & 0.2589\\
		\multicolumn{2}{l}{Ours} 
		& \textbf{33.2161} & \textbf{0.9203} & \textbf{0.1178} & \textbf{0.2186} \\
		\bottomrule
	\end{tabular}
	\caption{\textbf{Quantitative comparison against several baseline methods in terms of rendering accuracy.} Compared with NeRF, NeRF-T, Neural Volumes, AGI, and HVR, our approach achieves the best performance in all four metrics.}
	\Description{Fully described in the text.}
	\label{table:comparison}
\end{table}

\subsection{Novel Space-time Editing Results}
\begin{figure*}[t]
  \includegraphics[width=\linewidth]{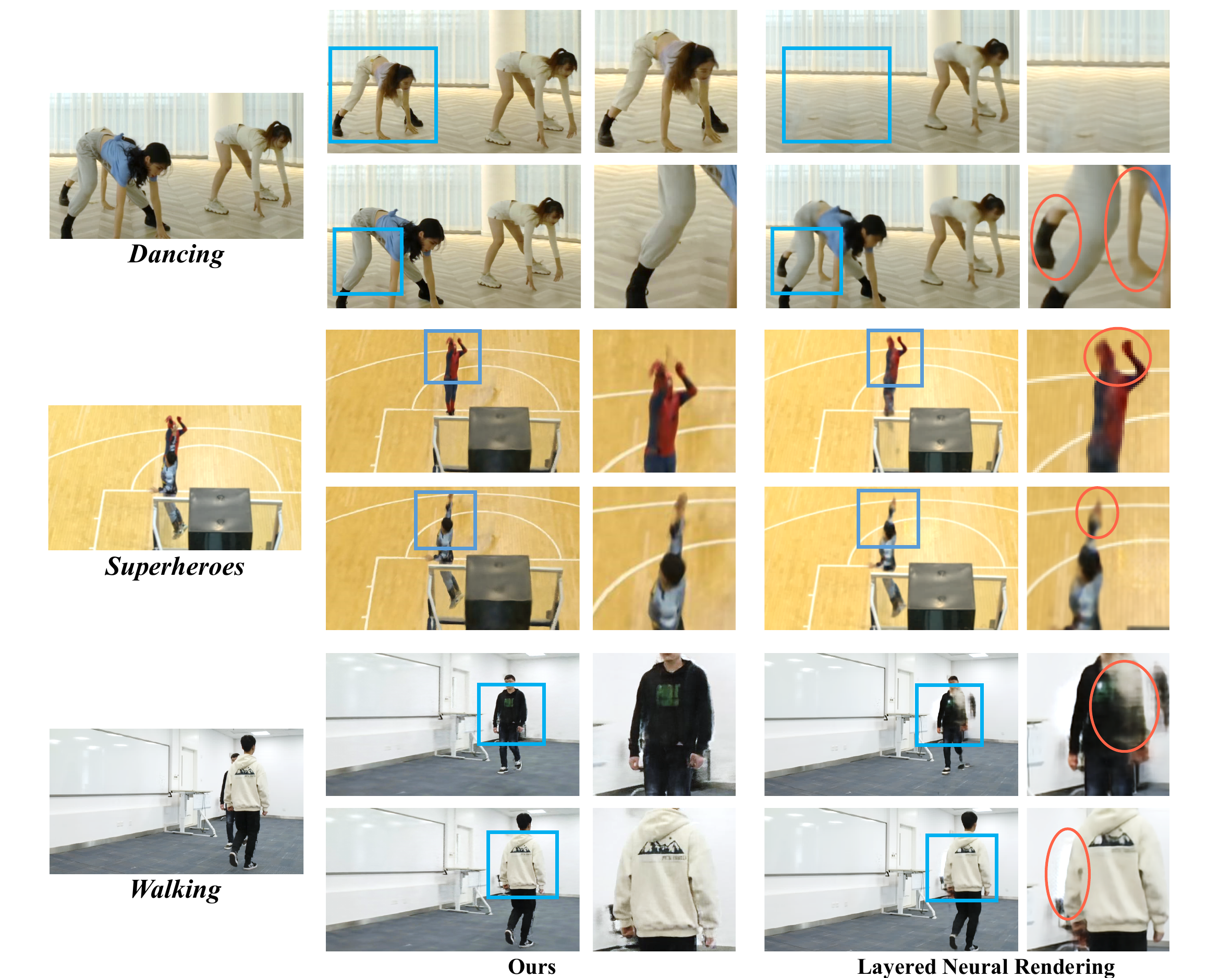}
  \caption{\textbf{Comparison in terms of layered reconstruction.} We show our layered reconstruction results and compare them with the baseline method, Layered Neural Renderings. For each sequence, the first and second rows illustrate the reconstruction result of the occluded and the front performer, respectively. We highlight the representative reconstruction results with blue 2D bounding boxes. Furthermore, we use orange ellipses as auxiliary descriptions.}
  \label{fig:layered_comparison}
  \Description{Fully described in the text.}
\end{figure*}

Our simple yet expressive editing functions achieve depth-aware and photo-realistic free-viewpoint video editing results. 
As demonstrated in Fig.~\ref{fig:gallery}, our approach generates editable free-viewpoint videos for complicated dynamic scenes in our dataset, e.g., dancing, playing basketball, or playing music. 
Fig.~\ref{fig:teaser} provides the representative results of our free-viewpoint videos with fancy editing effects. 
Note that we edit the position and rotation of the violinist on the top row and duplicate the two performers with various scales on the bottom row, respectively.
In Fig.~\ref{fig:occluded}, two performers pass by each other, and they are occluded by each other at a specific view for a short time. 
We demonstrate that our approach can successfully recover the occluded regions for the performers. Moreover, the corresponding comparison about such recovery against the Layer Neural Representation (LNR)~\cite{lu2020layered} is provided in Fig.~\ref{fig:layered_comparison}. 

In our supplementary video, we show the complete editing results for different scenes. Our ST-NeRF can achieve realistic editing for various layers represented as continuous functions with the consistency of space, which is hard to achieve by an image-based editing method.  
For \textit{Breaking}, we set keyframes for each dancer, adapting their actions to the beats of background music. Our approach shows the ability to retime different entities individually. 
For \textit{Taekwondo}, two actors perform the same action asynchronously, and we manually set keyframes for each layer aligning with the same global keyframes. Our edited free-viewpoint video achieves synchronous action at a novel time for them. 
For \textit{Musicians}, to obtain the desired layout of the free-viewpoint video, we shift the violinist closer to the pianist so that the layout is harmonious for the audience. 
For \textit{Superheroes}, the cameras were set far from two superheroes, so we firstly zoom in to focus on their one-on-one. Then freeze the time of spider-man when he is doing a jump shot, finally rendering a wide range free-viewpoint video to appreciate his act of shooting. Since our model can successfully decompose this scene into four layers: glass, batman, spider-man, and background, after hiding the glass, we can correctly render the front view of spider-man, which is occluded by the glass in the original viewpoint. 
For \textit{K-pop}, we designed a series of camera trajectories combining with retiming for some specific frames to achieve an artistic free-viewpoint video rendering result similar to a music video.

\begin{figure}[t]
	\includegraphics[width=\linewidth]{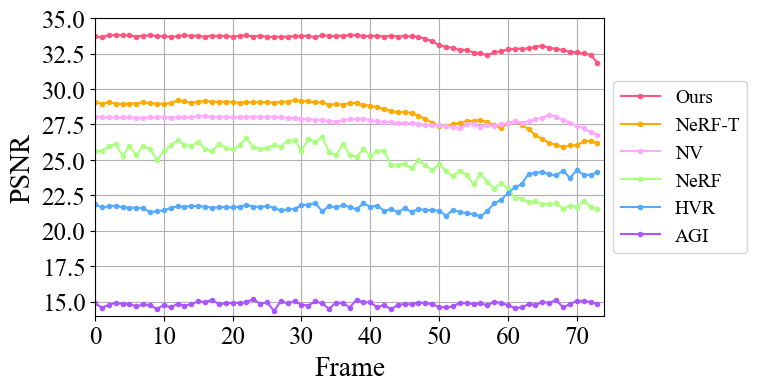}
	\caption{\textbf{Quantitative comparison against various baseline methods in terms of PSNR.} Our approach consistently achieves the highest PSNR for all the frames of the \textit{Walking} sequence.}
	\label{fig:curve1}
	\Description{Fully described in the text.}
\end{figure}

\begin{figure*}[th]
	\includegraphics[width=0.95\linewidth]{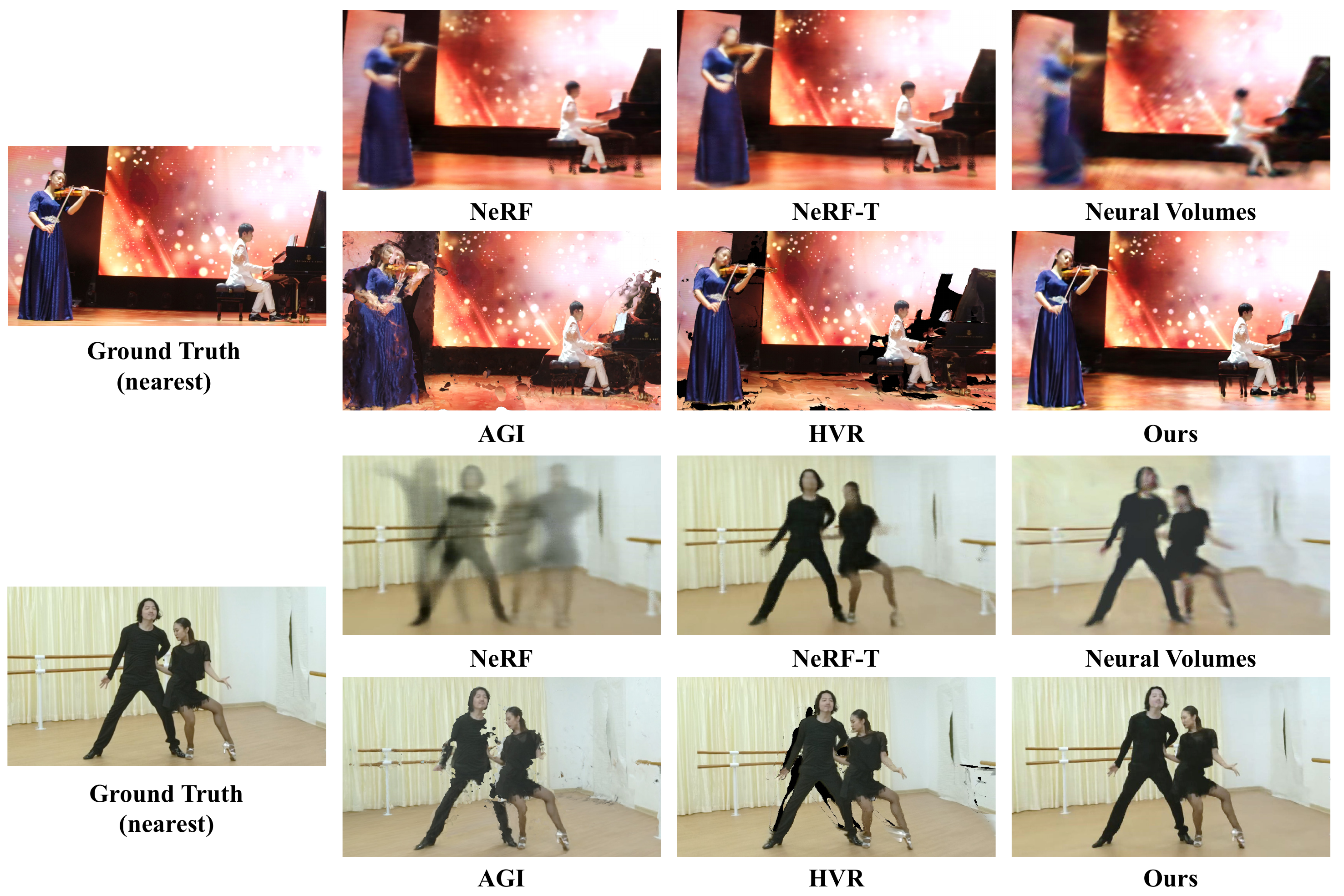}
	\caption{\textbf{Qualitative comparison with Neural Volumes, NeRF, NeRF-T, AGI and HVR.} Note that our approach generalizes the most photo-realistic and finer details.}
	\Description{Fully described in the text.}
	\label{fig:compare}
\end{figure*}

\subsection{Comparisons}
To the best of our knowledge, our approach is the first approach to generate editable free-viewpoint videos using a layered neural representation.
To demonstrate the overall performance of our approach, we compare to the existing free-viewpoint video methods based on neural rendering, including the voxel-based method \textbf{Neural Volumes}~\cite{NeuralVolumes}, the implicit method \textbf{NeRF}~\cite{mildenhall2020nerf} based on neural radiance field, and a variation of NeRF by natively adding time as input, denoted as \textbf{NeRF-T}.
Additionally, we compare our approach with a traditional method \cite{zitnick2004high}, which enables high-quality video-based rendering(\textbf{HVR}) of dynamic scenes based on segmentation-based stereo.
%
For a thorough comparison, we further compare against the traditional mesh-based modeling pipeline using the commercial software Agisoft PhotoScan~\cite{verhoeven2011taking}, denoted as \textbf{AGI}.
For a fair comparison, Neural Volumes, NeRF, and NeRF-T share the same training dataset as our approach, and we reconstruct the scene to obtain a textured mesh for every single frame in AGI from all the input viewpoints. 

%
For quantitative comparison, we adopt the peak signal-to-noise ratio (\textbf{PSNR}), structural similarity index (\textbf{SSIM}), mean absolute error (\textbf{MAE}), and Learned Perceptual Image Patch Similarity (\textbf{LPIPS}) \cite{zhang2018perceptual} as metrics to evaluate our rendering accuracy. 
Note that we calculate all the quantitative results in all the captured reference views. 
As shown in the Tab.~\ref{table:comparison}, our approach outperforms all the other methods in terms of PSNR, SSIM, and MAE, showing the effectiveness of our model to provide a realistic rendering of the complicated dynamic scenes. 
In Fig.~\ref{fig:curve1}, we further provide the numerical curve of PSNR for the whole \textit{Walking} sequence.
Note that we calculate the average PSNR in the 16 capture views for each timestamp. 
Our approach consistently achieves the highest PSNR for all the frames compared to other baselines.
For qualitative comparison, we show the novel view rendering results and the nearest input view in Fig.~\ref{fig:compare}.
NeRF can only handle the static scene, and its variation NeRF-T suffers from severe blur artifacts in the rendering results due to the challenging motions of various performers. 
Neural Volumes can provide more reasonable rendering results, but it still suffers from uncanny blur results due to the limited resolution of the voxel grid.
Moreover, AGI and HVR generate sharper rendering appearance results but are limited by the reconstruction accuracy, leading to severe artifacts in those missing regions, especially near the boundary. 
In contrast, our approach achieves the most vivid rendering result in terms of photo-realism and sharpness. 
Additionally, we generate more consistent rendering results from different views and timestamps without flickering with the space-time training in our approach.
These qualitative and quantitative comparisons above reveal the effectiveness of our method for better novel view synthesis for large-scale dynamic scenes.
Also, note that our approach enables various editing functions for fancy visual effects unseen in previous baselines.

We further compare our approach against the \textbf{Layered Neural Rendering}~\cite{lu2020layered} qualitatively to evaluate our performance for layered scene reconstruction and rendering.
As the baseline~\cite{lu2020layered} only requires a monocular video as a single input, for a fair comparison, we only generate the rendering results in the input view of Layered Neural Rendering.
As shown in Fig.~\ref{fig:layered_comparison}, Layered Neural Rendering~\cite{lu2020layered} fails to segment the different dynamic entities due to the severe inherent self-occlusion due to the single-view setting, though \cite{lu2020layered} gives comparable reconstruction results in the input view.
In contrast, our approach yields depth-aware and physically correct rendering of the two overlapped performers in the capture view. 
Such a qualitative comparison illustrates the effectiveness of our approach to encode the spatial and temporal information from our multi-view setting, which enables accurate decomposition and impressive rendering results for immersive free-view experiences.

\subsection{Ablation Study}
\begin{figure}[t]
	\includegraphics[width=\linewidth]{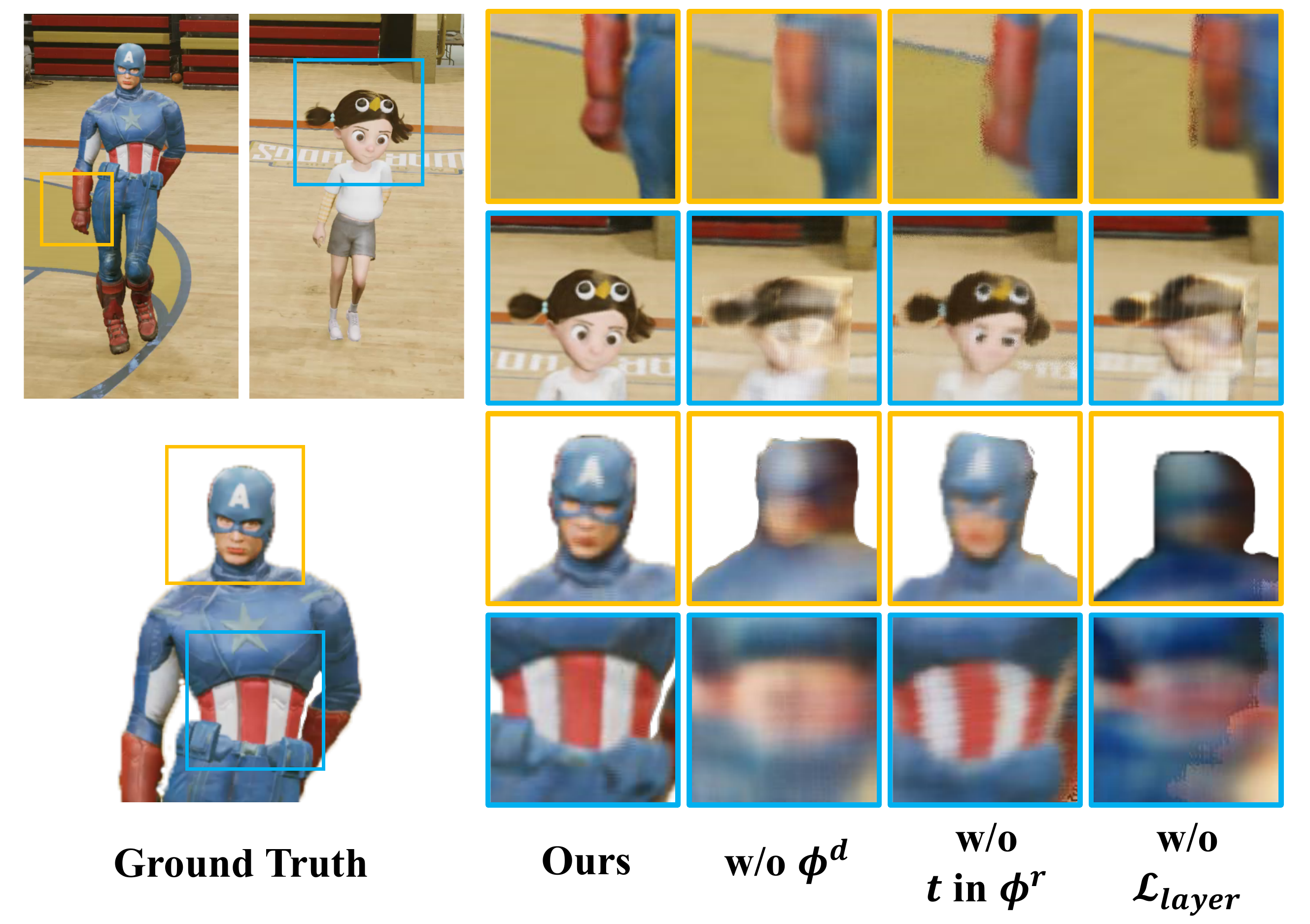}
	\caption{\textbf{Qualitative ablation study for our technical components.} This evaluation demonstrates the contribution and effectiveness of the three algorithmic components.}
	\label{fig:evaluation}
	\Description{Fully described in the text.}
\end{figure}

Here, we evaluate the performance of different components and loss terms in our approach.
Let \textbf{w.o $\phi^{d}$}, \textbf{w.o $t$ in $\phi^r$} and \textbf{w.o. $\mathcal{L}_{layer}$} denote the variations of our approach without the deformation net $\phi^d$, without inputting time $t$ into radiance net $\phi^r$ and without the layer-wise loss $\mathcal{L}_{layer}$, respectively.
As shown in Fig.~\ref{fig:evaluation}, without the deform module, our approach cannot handle dynamic entities in the scene, while the lack of inputting time $t$ into the radiance module leads to blurring rendering artifacts. 
Moreover, without layer loss, the training leads to a wrong decomposition result of the scene. 
In contrast, our complete approach achieves photo-realistic results with better decomposition for various entities.

\begin{figure}[t]
	\includegraphics[width=\linewidth]{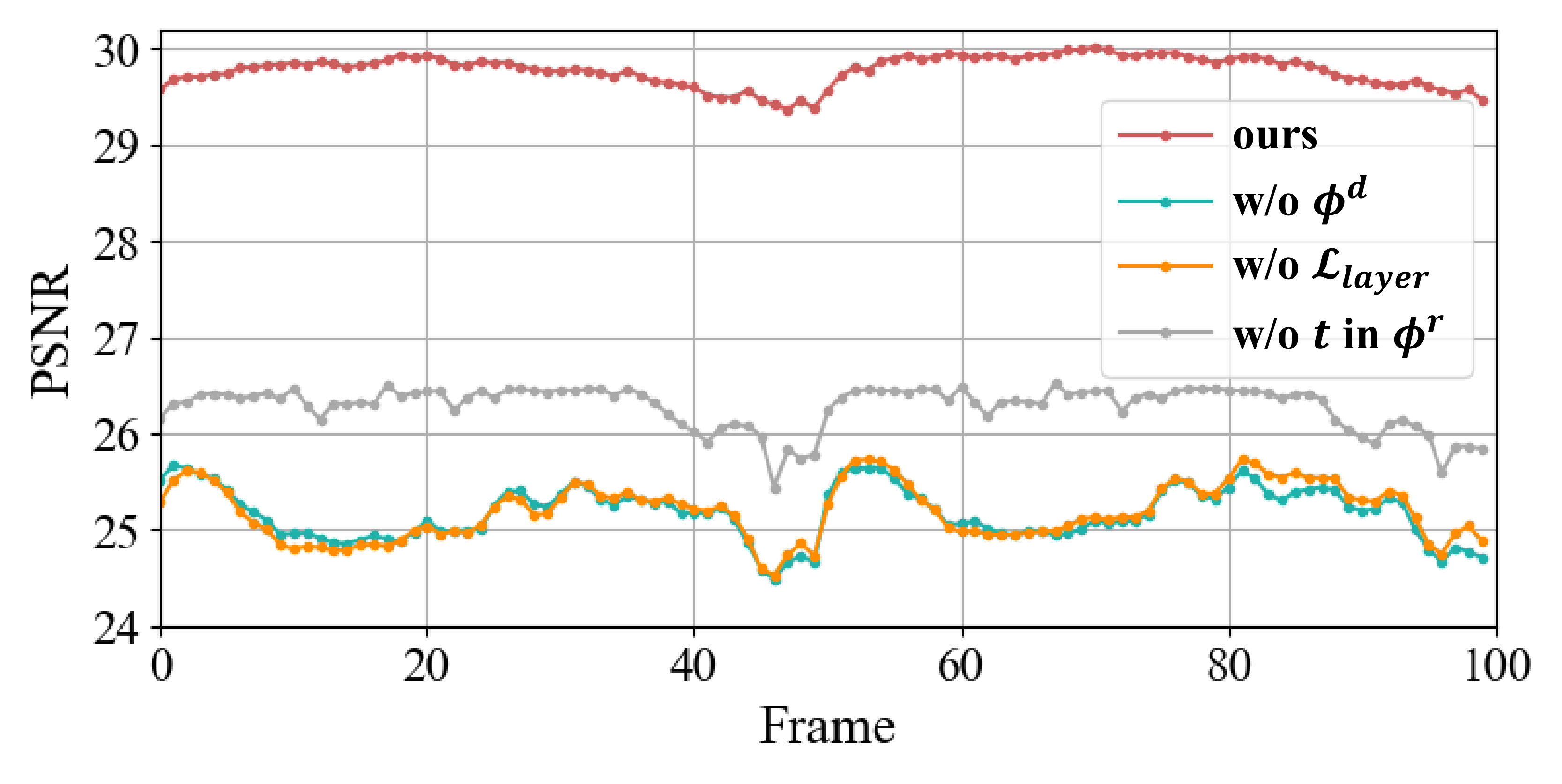}
	\caption{\textbf{PSNR curves of different variations.} This quantitative evaluation illustrates that our complete model consistently achieves the best performance across different frames.}
	\label{fig:curve2}
	\Description{Fully described in the text.}
\end{figure}

To further analyze variations of our approach, we utilize the same four metrics to evaluate the performances quantitatively.
We obtain the quantitative results in terms of PSNR, SSIM, MAE, and LPIPS for each variation by averaging the results of all the frames and views in our synthetic dataset. 
Besides, we compare against our variations in a held-out view and calculate the average PSNR in all the frames, denoted as \textbf{PSNR$_{test}$}. 
Our approach outperforms other variations in terms of all the metrics as shown in Tab.~\ref{table:ablation}.
Furthermore, as shown in Fig.~\ref{fig:curve2}, we compare the PSNR curves for these specific variations. 
Note that we calculate the value of PSNR for a single frame by averaging the views' results. 
Our complete approach consistently achieves the best result comparing with other variations.

\begin{figure*}[t]
	\includegraphics[width=\linewidth]{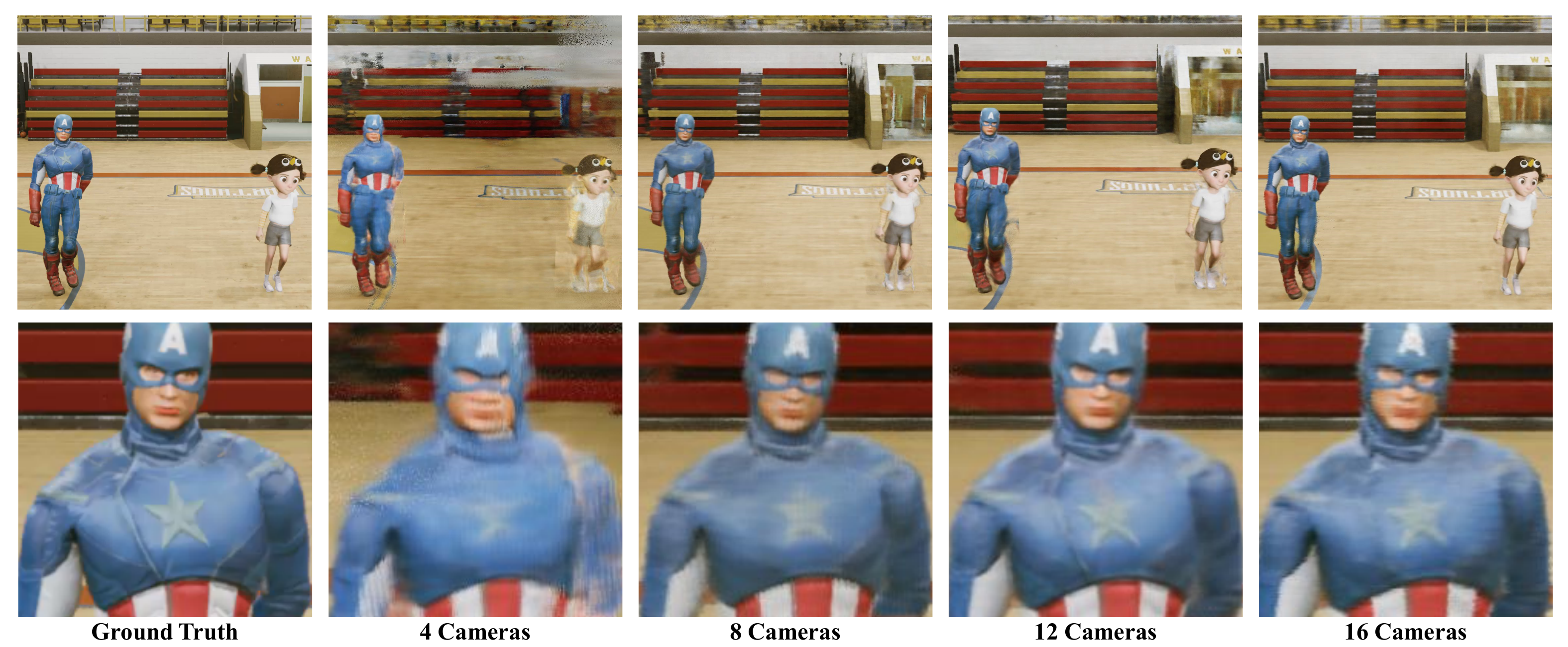}
	\caption{\textbf{Qualitative view ablation study.} This evaluation demonstrates the illustration of held-out view rendering results in different view numbers. Using a more significant view number gives a better result in the held-out view.}
	\label{fig:evaluation2}
	\Description{Fully described in the text.}
\end{figure*}

We further evaluate our model with different numbers of views. 
Specifically, there are a total of 17 cameras in our synthetic dataset, uniformly ranging from 0 to 160 degrees in a circle in an outside-in manner. 
Then, we evaluate the variations of our approach using 16, 12, 8, and 4 cameras for both input and training, respectively. 
To evaluate our model performance fairly, we utilize the central camera as the held-out view and calculate its corresponding PSNR, SSIM, MAE, LPIPS as quantitative metrics. 
We also evaluate the average PSNR in all the training views, denoted as \textbf{PSNR$_{train}$}.
As shown in Tab.~\ref{table:ablation2}, in the held-out view, the performance of our approach steadily goes down in terms of all the metrics when decreasing view number, while the model with less training views has slightly higher PSNR$_{train}$ due to over-fitting.
The corresponding qualitative results are provided in Fig.~\ref{fig:evaluation2}. 
When the number of cameras decreases to less than 8, the rendering result of the held-out view is getting worse, leading to severe artifacts, e.g., ghosting and wrong reconstruction of the 3D scene.
%
\begin{table}[tp]
\small
\begin{tabular}{@{}ccccccc@{}}

\toprule
\multicolumn{6}{c}{Ablation study of our model components} \\ \midrule
\multicolumn{1}{c}{Method} &  PSNR$_{test}$($\uparrow$) & PSNR($\uparrow$) & SSIM ($\uparrow$) & MAE ($\downarrow$) & LPIPS ($\downarrow$)\\
\multicolumn{1}{c}{w.o $\phi^{d}$} & 25.5461 & 25.6031      & 0.7878      & 0.0301  & 0.3962\\
\multicolumn{1}{c}{w.o $t$ in $\phi^r$}& 26.0818 & 26.5394      & 0.8015      & 0.0272  & 0.3800\\
\multicolumn{1}{c}{w.o. $\mathcal{L}_{layer}$}& 25.1343 & 25.7336      & 0.8016      & 0.0296  & 0.3960\\
\multicolumn{1}{c}{Ours} & \textbf{29.9091} & \textbf{30.0502} & \textbf{0.8566} & \textbf{0.0187}     & \textbf{0.2329}\\ \bottomrule
\end{tabular}
\caption{\textbf{Quantitative model ablation study.} $\uparrow$ means larger is better, while $\downarrow$ means smaller is better. Our complete pipeline outperforms other variations in terms of four metrics.}
\label{table:ablation}
\Description{Fully described in the text.}
\end{table}

\subsection{Limitations and Discussions}

\begin{table}[t]
	\small
	\begin{tabular}{@{}cccccc@{}}
		\toprule
		\multicolumn{6}{c}{Ablation study of number of views} \\ \midrule
		\# of views  & PSNR$_{train}$($\uparrow$) & PSNR($\uparrow$) & SSIM($\uparrow$)     & MAE($\downarrow$)      & LPIPS($\downarrow$)    \\
		4            &\textbf{ 30.9227}  & 15.9286 & 0.7273  & 0.1175  & 0.5321  \\
		8            & 28.4074  & 22.1213 & 0.8512  & 0.0424  & 0.2347  \\
		12           & 28.1643  & 23.2100 & 0.8580  & 0.0382  & 0.2230 \\
		16           & 27.9974  & \textbf{26.3877} &\textbf{ 0.8866}  &\textbf{ 0.0261}  & \textbf{0.1940}  \\ \bottomrule
	\end{tabular}
	\caption{\textbf{Quantitative view number ablation study.} $\uparrow$ means larger is better, while $\downarrow$ means smaller is better. Merely inputting four views during training makes the network overfit and gives a higher training view PSNR. On the other hand, inputting sixteen views during training outperforms in held-out view metrics, showing better novel view rendering results.}
	\label{table:ablation2}
	\Description{Fully described in the text.}
\end{table}

We have demonstrated the compelling capability of editable free-viewpoint video generation with a variety of space-time editing functions in a photo-realistic manner unseen before.
Nevertheless, as the first trial to combine such editable free-viewpoint video with a layered neural representation, our approach is subject to some limitations.

First, our scene parsing stage relies on the color difference for label map tracking and may fail when handling dynamic entities with a similar appearance. Also, because our scene sensing scheme relies on a human segmentation algorithm~\cite{wang2019fast}, we only show our results focusing on dynamic humans. It would be interesting to consider each object as a layer enabling a more editable dynamic scene.
Furthermore, our approach cannot handle those extremely challenging scenarios with severely occluded entities where the bounding box tracking fails.
Besides, the tracked bounding box serves as the spatial anchor when training our ST-NeRF for an individual.
Thus, the rich information outside the bounding box cannot be obtained by the network, leading to a worse view-dependent effect, especially for those light-changing scenarios.
Such un-modeled regions will be learned into the background layer, leading to 3D ghosting artifacts when editing the related neural layers. 
However, such a case can be easily fixed by manually correcting the corresponding bounding box. 
%
Currently, our approach still relies on 16 cameras to provide a wide range of free-viewing. 
It is a promising direction to reduce the camera number by adopting more data-driven scene modeling strategies or utilizing a pre-scanned static background as an initial proxy.
Furthermore, we only adopt some basic spatial and temporal editing functions in our pipeline, which already provides promising fancy visual effects with high realism, while the functions do not support non-rigid manipulation or slow-motion effects. Finally, we use a simple nearest interpolation scheme when generating new timestamps for each layer.
In the future, we plan to explore more non-rigid editing functions in the same framework with layered neural representation. 
It is promising to encode more human motion prior for such non-rigid effects, e.g., using the human template model SMPL~\cite{loper2015smpl} as a spatial and temporal anchor.
It's also interesting to model the illumination and lighting effect in a large-scale dynamic scene to enable more controllable disentanglement or re-lightable editing.

\section{CONCLUSION}
We have presented the first approach to generate high-quality editable free-viewpoint videos of large-scale dynamic scenes from relatively sparse 16 RGB cameras.
Our novel pipeline enables a variety of photo-realistic space-time visual editing effects while still supporting wide-range free viewing.
The core of our approach is a new layered neural representation where each layer learned the spatially and temporally consistent correlation between an individual and the dynamic scene to support various editing functions. 
Our neural representation enables the disentanglement of location, deformation, and the appearance of various dynamic entities.
Our deform module encodes the temporal motion robustly, while our object-aware volume rendering scheme enables the re-assembling of all the neural layers.
Our neural editing enables explicit spatial and temporal manipulations of various dynamic entities in the scene while maintaining high realism and supporting wide-range free viewing.
Extensive experimental results demonstrate the effectiveness of our approach for editable free-viewpoint generation, which compares favorably to the state-of-the-art.
We believe that our approach renews the presence of free-viewpoint videos with more natural and controllable viewing ability, serving as a critical step for editable novel view synthesis, with many potential applications for fancy visual effects in VR/AR, gaming, filming, or entertainment.

\begin{acks}


The authors would like to thank Xin Chen, Dali Gao from DGene Digital Technology Co., Ltd. for processing the dataset and figures. Besides, we thank Boyuan Zhang from ShanghaiTech University for producing a supplementary video and  Yingwenqi Jiang, Wenman Hu, and Jiajie Yang for collecting the dataset as performers.
This work was supported by NSFC programs (61976138, 61977047), the National Key Research and Development Program (2018YFB2100500), STCSM (2015F0203-000-06) and SHMEC (2019-01-07-00-01-E00003).

\end{acks}
\balance
\bibliographystyle{ACM-Reference-Format}
\bibliography{ref}

\end{document}